\documentclass[10pt,twocolumn,letterpaper]{article}

\usepackage[pagenumbers]{cvpr} 

\definecolor{cvprblue}{rgb}{0.21,0.49,0.74}
\usepackage[pagebackref,breaklinks,colorlinks,allcolors=cvprblue]{hyperref}

\usepackage{amssymb}
\usepackage{makecell}
\usepackage{textcomp}
\usepackage{tabularx}
\usepackage{multirow}
\newcommand{\mytilde}{\raisebox{0.5ex}{\texttildelow}}

\newcommand{\ignore}[1]{}

\def\confName{CVPR}
\def\confYear{2026}

\title{
\mbox{}\\[-14mm]
Ego-1K -- A Large-Scale Multiview Video Dataset for Egocentric Vision
}

\author{
Jae Yong Lee      \hspace{7mm}
Daniel Scharstein \hspace{7mm}
Akash Bapat       \hspace{7mm}
Hao Hu            \hspace{7mm}
Andrew Fu\\
Haoru Zhao      \hspace{7.78mm}
Paul Sammut     \hspace{7.78mm}
Xiang Li        \hspace{7.78mm}
Stephen Jeapes  \hspace{7.78mm}
Anik Gupta\\
Lior David         \hspace{11.3mm}
Saketh Madhuvarasu \hspace{11.3mm}
Jay Girish Joshi   \hspace{11.3mm}
Jason Wither\\[1mm]
Meta Reality Labs
}

\begin{document}
\twocolumn[{%
\renewcommand\twocolumn[1][]{#1}%
\maketitle
\centering

\includegraphics[width=0.95\linewidth]{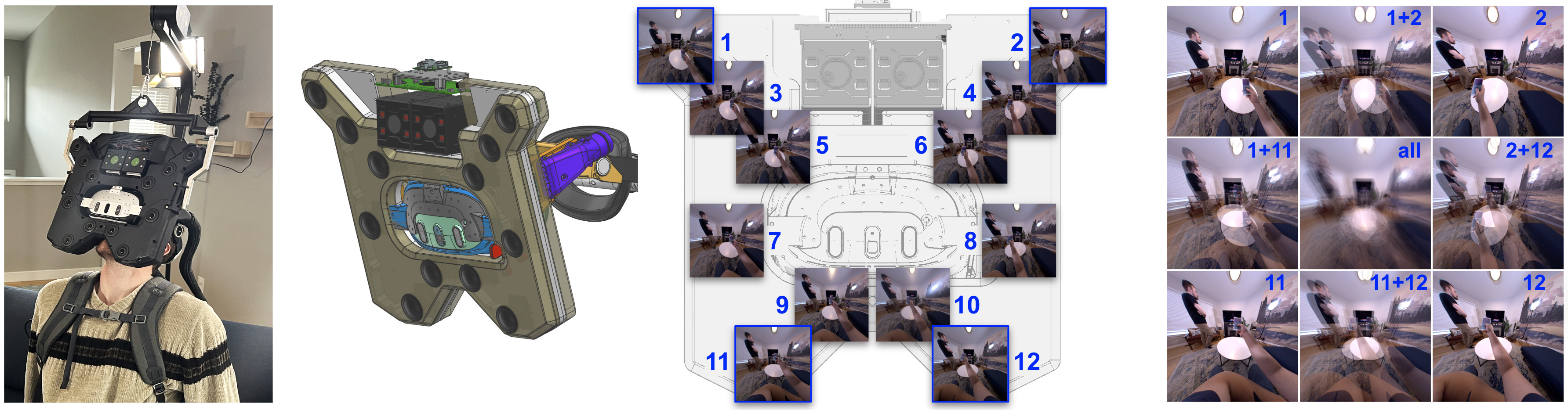}

\captionof{figure}{
Left: photo and rendering of our multi-camera rig integrating 12 global-shutter RGB fisheye cameras and a Quest 3 headset with 4 forward-facing cameras.  All cameras are synchronized, enabling the capture of dynamic egocentric multiview videos at 60 Hz. Middle: a sample frame from a dynamic scene, captured by the 12 rig cameras; each horizontal pair is stereo-rectified.  Right: overlays of the 4 corner views visualizing the disparity range; the average of all 12 views is shown in the center.
\vspace{1.5em}
}
\label{fig:teaser}
}]

\begin{abstract}

We present Ego-1K, a large-scale collection of time-synchronized egocentric multiview videos designed to advance neural 3D video synthesis and dynamic scene understanding.  The dataset contains nearly 1,000 short egocentric videos captured with a custom rig with 12 synchronized cameras surrounding a 4-camera VR headset worn by the user.  Scene content focuses on hand motions and hand-object interactions in different settings.  We describe rig design, data processing, and calibration.  Our dataset enables new ways to benchmark egocentric scene reconstruction methods, an important research area as smart glasses with multiple cameras become omnipresent.  Our experiments demonstrate that our dataset presents unique challenges for existing 3D and 4D novel view synthesis methods due to large disparities and image motion caused by close dynamic objects and rig egomotion.  Our dataset supports future research in this challenging domain.
It is available at 
\url{https://huggingface.co/datasets/facebook/ego-1k}.

\end{abstract}

    
\section{Introduction}
\label{sec:intro}

Mixed-reality devices and egocentric world modeling demand photorealistic 4D reconstruction from the wearer’s point of view. Yet, despite the rapid progress in neural novel view synthesis (NVS) and dynamic radiance field methods, there is no large-scale dataset that provides synchronized, multiview egocentric video of real, dynamic scenes. Existing NVS datasets are typically exocentric or monocular and focus on static scenes, while egocentric datasets prioritize activity recognition with monocular or stereo views, lacking the synchronized multiview imagery needed to drive and benchmark egocentric 4D reconstruction.  

Combining 4D reconstruction and egocentric vision presents compelling use cases, from remote presence to spatial reasoning and robotics. To support this novel research area, we introduce Ego‑1K, a dataset of 956 short egocentric recordings captured with a custom head-mounted rig that integrates 12 global-shutter RGB fisheye cameras surrounding a Quest 3 headset with 4 cameras (Fig.~\ref{fig:teaser}). All 16 cameras are hardware-synchronized at 60 Hz, enabling dynamic egocentric multiview capture with precise calibration and shared timestamps. The dataset emphasizes near-field hand-object interactions (HOI) with forward-facing coverage, posing unique challenges for reconstruction due to fast motion, frequent occlusions, and extreme disparities. We release both a ``raw'' version containing all sensor streams and a ``research'' version consisting of the 12 rig cameras dewarped into 6 rectified stereo pairs for easy processing; the research version is used in our experiments.

Our dataset enables new ways to benchmark egocentric scene reconstruction methods.  For stereo, we propose to evaluate pairwise consistency by warping disparity maps from different rig pairs into a chosen target pair and measuring agreement.  For novel view synthesis, we propose evaluating static per-frame 3D Gaussian splatting (3DGS) and 4D dynamic models using a train–test split where two target views are held out and the remaining ten rig views are used for training. Our experiments demonstrate that our dataset presents unique challenges for existing 3D and 4D NVS methods, which are ill-equipped to handle the combination of ego motion, near-range hand motion, large disparities, and frequent occlusions.  However, we also demonstrate that performance can be improved dramatically via depth guidance with current stereo foundation models.

Our main contributions are as follows. We introduce Ego-1K, a large-scale dataset of nearly 1K short egocentric videos of real, dynamic scenes, captured with a unique rig of 12+4 hardware-synchronized cameras.  Our dataset fills a critical gap by jointly achieving egocentric perspective, high camera count, and large scale.  It enables benchmarking of 3D video synthesis and dynamic novel view synthesis in complex real-world environments at the intersection of multiview stereo and egocentric 4D synthesis. In addition, we propose new evaluation protocols, demonstrate that existing dynamic NVS approaches fail under these challenging conditions, and that they can be improved by leveraging fused stereo depth as an additional prior.

\begin{table*}[t]
\centering
{
\footnotesize
\begin{tabular}{lc@{~~}c@{~~}c@{~~~}c@{~~~}c@{~~~}l@{~~~}l@{~~~}ll}
\toprule
\makecell[l]{Dataset\\~} & 
\makecell{Multi\\view} & 
\makecell{Ego-\\centric} &
\makecell{Large-\\scale} &
\makecell{\# Ego\\cams} & 
\makecell{\# Exo\\cams} &
\makecell[l]{\# Videos/frames\\~} &
\makecell[l]{Real/\\synth} & 
\makecell[l]{Interaction\\horizon} & 
\makecell[l]{Core benchmark task\\(geometry vs.~semantics)} \\
\midrule

\multicolumn{10}{l}{\hspace{-1mm}\textbf{Neural 3D Video Synthesis Datasets}} \\
NSFF \cite{nsff_li_cvpr21} &
-- & --  & -- & 0 & 1 & 8 videos & Real & Short / dynamics & Geom.~/ dynamic NVS \\
HyperNeRF \cite{hypernerf_park_tog21} &
-- & -- & -- & 0 & 1 & Few scenes & Real & Short / dynamics & Geom.~/ non-rigid NVS \\
DNeRF \cite{dnerf_pumarola_cvpr21} &
\checkmark & -- & -- & 0 & varies & Few scenes & Synth & Short / dynamics & Geom.~/ non-rigid NVS \\
Neural 3D Video \cite{neural3dvideo_li_cvpr22} &
\checkmark & -- & -- & 0 & 18 & 6 indoor scenes & Real & Short / dynamics & Geom.~/ multiview NVS \\
DiVA360 \cite{diva360_lu_cvpr24} &
\checkmark & -- & (\checkmark) & 0 & 53 & 54 videos & Real & Med.~/ dynamics & Geom.~/ dome NVS \\
\midrule

\multicolumn{10}{l}{\hspace{-1mm}\textbf{Egocentric Vision Datasets}} \\
Ego4D \cite{ego4d_grauman_cvpr22} &
-- & \checkmark & \checkmark & 1--3 & 0 & 20k+ videos, 3.7k hrs  & Real & Long / activities & Sem.~/ recognition \\
EPIC-KITCHENS \cite{epickitchens_damen_eccv18, epickitchens100_damen_ijcv22} &
-- & \checkmark & \checkmark & 1 & 0 & 100 hrs & Real & Long / activities & Sem.~/ recognition \\
HoloAssist \cite{holoassist_wang_iccv23} &
-- & \checkmark & \checkmark & 1 & 0 & Large-scale & Real & Med.~/ activities & Sem.~/ recognition \\
H2O \cite{h2o_kwon_iccv21} &
-- & \checkmark & \checkmark & 1 & 4 & 572k images & Real & Short / activities & Sem.~/ HOI recognition \\
HOI4D \cite{hoi4d_liu_cvpr22} &
-- & \checkmark & \checkmark & 1 & 0 & 4k videos, 2.4M images & Real & Short / dynamics & Sem.~/ HOI segmentation \\
ARCTIC \cite{arctic_fan_cvpr23} &
-- & \checkmark & \checkmark & 1 & 8 & 2.1M images & Real & Short / dynamics & Geom.~/ hand reconstr. \\
EgoObjects \cite{egoobjects_zhu_iccv23} &
-- & \checkmark & \checkmark & 1 & 0 & 9k+ videos & Real & Short / objects & Sem.~/ object detection \\
EgoPoints \cite{egopoints_darkhalil_wacv25} &
-- & \checkmark & \checkmark & 1 & 0 & Large-scale & Real & Short / dynamics & Geom.~/ point tracking \\
\midrule

\multicolumn{10}{l}{\hspace{-1mm}\textbf{Multiview Egocentric Datasets}} \\
EgoExo4D \cite{egoexo4d_grauman_cvpr24} &
(\checkmark) & \checkmark & \checkmark & 3 & 4--5 & 5k videos, 1,286 hrs & Real & Long / activities & Sem.~/ recog.~+ pose \\
EgoHumans \cite{egohumans_khirodkar_iccv23} &
(\checkmark) & \checkmark & (\checkmark) & 2--6* & 8--15 & 7 scenes, 125k images & Real & Med.~/ dynamics & Geom.~/ 3D tracking \\
EgoSim \cite{egosim_hollidt_neurips24} & 
(\checkmark) & \checkmark & \checkmark & 6** & 0 & 5h real + 100h synth & Mixed & Med.~/ activities & Geom.~/ human pose \\
HD-EPIC \cite{hdepic_perrett_cvpr25} &
(\checkmark)  & \checkmark & \checkmark & 3 & 0 & 156 videos, 41 hrs & Real & Long / activities & Sem.~/ recognition \\
HOT3D \cite{hot3d_banerjee_cvpr25} &
(\checkmark) & \checkmark & \checkmark & 2--3 & 0 & 198 Aria, 226 Quest & Real & Short / dynamics & Geom.~/ pose tracking \\
\bf Ego-1K (ours) & \bf\checkmark & \bf\checkmark & \bf\checkmark & \bf 12+4 & \bf 0 & \bf 956 videos, 514k frames & \bf Real & \bf Short / dynamics & \bf Geom.~/ HOI dyn.~NVS \\
\bottomrule
&&&\multicolumn{7}{l}{~~~~~~~~~ *only 1 egocentric view per subject ~~~~~ ** only 1 egocentric view from the user's head}
\end{tabular}
}
\caption{Comparison of existing datasets for dynamic 3D video synthesis and multiview egocentric vision.  Our dataset is the only one that provides synchronized egocentric multiview captures, with all 12+4 cameras following the user's head motion.}
\label{table:datasets}
\end{table*}

\section{Related Work}
\label{sec:related}

Stereo datasets have played a pivotal role in advancing the field of 3D reconstruction, providing the foundational data necessary for developing and benchmarking algorithms that infer scene geometry from multiple viewpoints. These datasets have driven progress in both classical stereo matching and learning-based approaches, improving depth estimation, scene understanding, and visual SLAM. 
Diverse datasets such as Middlebury~\cite{scharstein_ijcv02}, KITTI~\cite{geiger_cvpr12}, Sintel~\cite{sintel_butler_eccv12}, DTU~\cite{jensen_cvpr14}, ETH3D~\cite{eth3d_schoeps_cvpr17}, and Tanks and Temples~\cite{knapitsch_tog17}
have been instrumental in benchmarking and advancing stereo and multiview stereo algorithms. Subsequent efforts have focused on large-scale synthetic datasets and foundation models for stereo, further expanding the scope and generalization of 3D reconstruction methods \cite{crestereo_li_cvpr22, foundationstereo_wen_cvpr25}.

More recently, the research community has shifted focus toward neural novel view synthesis, which aims to generate photorealistic images from unseen viewpoints. Early work in this area concentrated on static scenes, leveraging multiview images and neural representations to synthesize new perspectives with high fidelity 
\cite{nerf_mildenhall_eccv20, nerfwild_martinbrualla_cvpr21, plenoctrees_yu_iccv21, mipnerf_barron_iccv21, mvsnerf_chen_iccv21}.
Building on these successes, subsequent research has extended these methods to dynamic scenes, tackling the additional challenges posed by temporal changes and non-rigid motion
\cite{nsff_li_cvpr21, dnerf_pumarola_cvpr21, nerfies_park_iccv21, videonerf_xian_cvpr21, kplanes_fridovich_cvpr23, spacetime_li_cvpr24}.
 
Parallel to these developments, egocentric video analysis has emerged as an active area of study, driven by the proliferation of wearable cameras and the growing interest in understanding first-person experiences. Egocentric datasets have enabled progress in activity recognition, object interaction, and social understanding from a personal viewpoint \cite{epickitchens_damen_eccv18, egtea_li_eccv18, ego4d_grauman_cvpr22, hoi4d_liu_cvpr22}.
Recent benchmarks such as Ego4D~\cite{ego4d_grauman_cvpr22} and EFM3D~\cite{efm3d_straub24} have further advanced the field by providing large-scale, diverse, and richly annotated egocentric video data, supporting a wide range of research in perception, action understanding, and 3D scene analysis.

Despite these advances, there remains a critical gap: no existing dataset provides dense synchronized multiview video captured from an egocentric perspective. Such a resource is essential for bridging the domains of 3D reconstruction, NVS, and egocentric analysis, and unlocks new opportunities for research at their intersection. Our work addresses this gap by introducing Ego-1K, a new synchronized egocentric multiview dataset designed to facilitate progress across these rapidly evolving fields. Table~\ref{table:datasets}  provides a structured comparison of Ego-1K with existing datasets, including two axes that cut across all three lines of work: the interaction horizon (duration and focus) and the core benchmark task (geometry vs. semantics).

\subsubsection*{Neural 3D video synthesis datasets}

Early datasets focus on monocular dynamic videos: NSFF~\cite{nsff_li_cvpr21} uses monocular dynamic videos~\cite{yoon_cvpr20} to learn spacetime view synthesis, and HyperNeRF~\cite{hypernerf_park_tog21} uses monocular selfie-style videos to learn a dynamic deformation field per time.  D-NeRF~\cite{dnerf_pumarola_cvpr21} provides synthetic multiview dynamic objects in dome-capture style.  Neural 3D video synthesis~\cite{neural3dvideo_li_cvpr22} provides 21 GoPro videos of 6 indoor scenes, and DiVA-360~\cite{diva360_lu_cvpr24} provides dome-style captures from 53 time-synchronized cameras.  All these datasets lack egocentric perspectives and are limited in scale.

\subsubsection*{Egocentric vision datasets}

Large-scale datasets like Ego4D~\cite{ego4d_grauman_cvpr22} (3,670+ hours), EPIC-KITCHENS~\cite{epickitchens_damen_eccv18, epickitchens100_damen_ijcv22} (100 hours), and HoloAssist~\cite{holoassist_wang_iccv23} provide extensive egocentric videos but focus on activity recognition with monocular or stereo capture. While valuable for first-person vision research, they lack multiple views for dynamic 3D reconstruction. Similarly, EgoObjects~\cite{egoobjects_zhu_iccv23} is a large-scale monocular egocentric video dataset with diverse object labels, and EgoPoints~\cite{egopoints_darkhalil_wacv25} is a large-scale dataset for egocentric point tracking.

Several large-scale datasets focus on hand-object interaction (HOI): H2O~\cite{h2o_kwon_iccv21} is a benchmark designed for egocentric HOI recognition, HOI4D~\cite{hoi4d_liu_cvpr22} targets semantic and action segmentation and pose tracking, and ARCTIC~\cite{arctic_fan_cvpr23} focuses on manipulation of articulated objects.  However, they all provide only a single egocentric view.

\subsubsection*{Multiview egocentric datasets}
 
Existing datasets combining egocentric and multiview captures typically only feature a small number of synchronized egocentric cameras or have limited scale. Ego-Exo4D~\cite{egoexo4d_grauman_cvpr24} offers large-scale ego-exo data but uses only 3 cameras per setup for skill demonstration. EgoHumans~\cite{egohumans_khirodkar_iccv23} captures multiple views for human pose estimation but only features 7 scenes with limited realism. EgoSim~\cite{egosim_hollidt_neurips24} provides 6 GoPro recordings from human joint locations for human pose estimation, but only a single egocentric view per subject.
Other recent datasets like HD-EPIC~\cite{hdepic_perrett_cvpr25} and HOT3D~\cite{hot3d_banerjee_cvpr25} leverage modern head-mounted devices like Project Aria and Quest 3 to capture synchronized multiview egocentric videos. While HD-EPIC provides highly detailed, unscripted recordings of kitchen activities, and HOT3D provides a large collection for 3D hand and object tracking, both are limited to 2--3 egocentric views. In contrast, our dataset consists of 4 headset views plus 12 surrounding views, all synchronized.

\subsubsection*{3DGS with geometry priors}

As we show in Section~\ref{sec:experiments}, our dataset poses additional challenges of large disparities and fast image motion, causing problems for existing dynamic NVS methods \cite{splatfields_mihajlovic_eccv24, spacetime_li_cvpr24}.  Current stereo models, however, can often handle these challenges, and thus we propose to use stereo depth as a geometric prior by initializing splats from surfaces reconstructed via depth map fusion for each frame (Section~\ref{sec:geom_priors}).

Existing 3DGS methods employing geometry priors typically focus on regularization \cite{stereogs_safadoust_bmvc24, dro_chung_cvprw24, gsslam_matsuki_cvpr24}.  Regularization can improve NVS fidelity when initialized with a sparse point cloud, but this does not apply to our dense stereo initialization.  Several recent methods share a similar initialization scheme as ours: DN-Splatter~\cite{dnsplatter_turkulainen_wacv25} uses fused point clouds, and EDGS~\cite{edgs_kotovenko_arxiv25} uses dense correspondences for initialization of 3D Gaussians. DepthSplat~\cite{depthsplat_xu_cvpr25} proposes jointly training depth estimators and Gaussian splatting for feed-forward models and relies purely on learned depth, while our approach uses stereo-based TSDF fusion to obtain metrically accurate, watertight geometry that serves as a more reliable structure for NVS.

\section{Dataset}
\label{sec:dataset}

\subsubsection*{Multi-camera rig}

We collect synchronized multiview videos from a head-mounted rig integrating a Quest 3 headset, 12 fisheye cameras surrounding the headset, and two iToF sensors (Fig.~\ref{fig:teaser}).

The 12 external rig cameras are 8 MP (2848 × 2848) global shutter RGB sensors, fitted with 190° diagonal FOV f2.8 lenses. The sensors are cropped to 6 MP (2448 × 2448) to allow streaming at 60 FPS in 8-bit raw Bayer format over a USB 3.1 (5Gbps) connection to a backpack-mounted computer. The computer utilizes two 8-port USB host bus adapters to ingest the camera data and store it temporarily in RAM before saving to disk.  The rig also contains two Lucid Helios2 Wide iToF sensors (640w × 480h), alternately capturing at 30 FPS, time-synchronized with the cameras.
We include the iToF streams in the raw dataset but do not use them in our experiments due to unreliable depth estimates in the presence of motion and phase ambiguity.

The Meta Quest 3 mixed-reality headset enables users to experience immersive VR content while also supporting passthrough that blends digital content with the real world. The Quest 3 features two forward-facing rolling-shutter RGB cameras (2328w × 1748h) capturing at 60 FPS and two forward-facing global-shutter grayscale SLAM cameras (512w × 640h) capturing at 30 FPS. We do not capture the two side-facing SLAM cameras since their view is occluded by the rig. The headset’s VIO system provides rig poses derived from the SLAM cameras and an IMU; our raw dataset also includes the raw 6-DOF IMU signals at 800 Hz. Due to the rig’s weight (approximately 6 kg), we use a backpack-mounted crane to support most of it while allowing natural head motion.

\begin{figure}[t]
    \centering
    \includegraphics[width=\linewidth]{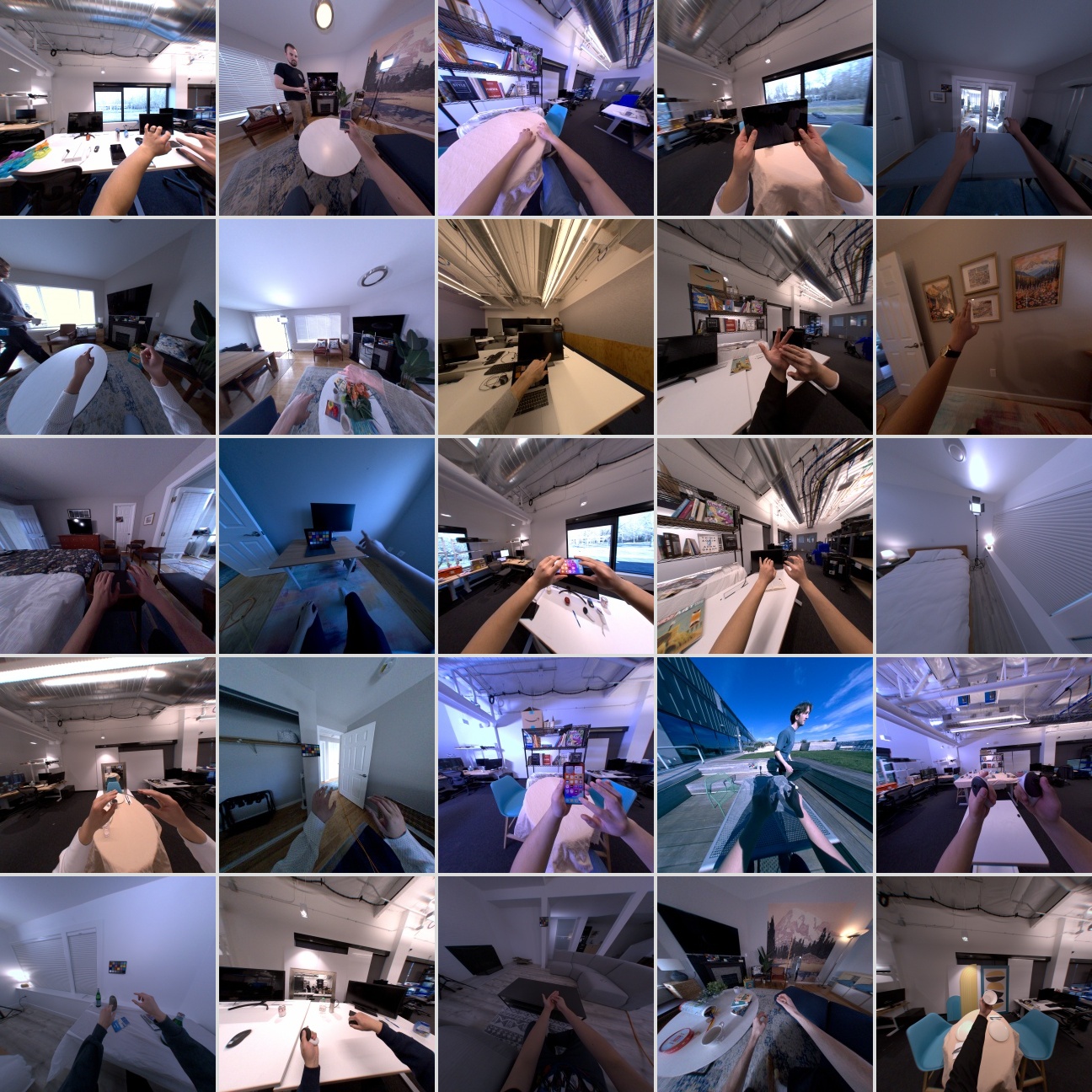}
    \caption{Sample frames from our dataset, illustrating the range of settings and hand motions.}
    \label{fig:dataset_samples}
\end{figure}

\subsubsection*{Scene content}

We collected 999 recordings with five operators wearing the rig 
across a range of environments, activities, and lighting conditions.
We focus on hand-object interaction: the operators perform various gestures, simulate typing, and manipulate different objects.  In some scenes, the operators also appear as bystanders.
Table~\ref{tab:dataset_properties} summarizes the different diversity axes and range of properties of our dataset.  We provide some of these properties in the form of text labels in the metadata accompanying the recordings.
We manually checked all videos and discarded 43 problematic recordings with dark images or an insufficient number of frames.  The remaining 956 recordings form our main dataset.  Fig.~\ref{fig:dataset_samples} shows sample frames from one of the rig cameras.

\subsubsection*{Data capture and processing}

When recording with the rig, we generate a substantial amount of data: 12+2 RGB video streams at 60 Hz, two SLAM streams at 30 Hz, two iToF streams at 30 Hz, and IMU data at 800 Hz.  This data is streamed to DDR RAM on the backpack PC and later transferred to an NVMe drive for non-volatile storage, generating about 15 GB/s of uncompressed data. For our dataset, we recorded videos of 8--10 seconds, typically 450--550 frames per camera.

To time-align the 12+4 cameras, a wireless synchronizer sends an LTC timestamp to both the Quest headset streams and the 12 rig cameras.  A timing controller communicates the timestamp for the triggered frames via MQTT to the ROS2 capture software running on the PC. This shared timestamp allows subsequent frame alignment between the VRS file containing the 12 rig cameras and the VRS file containing Quest 3 data, which is collected on the headset.

After uploading to a server, the rig VRS and the headset VRS are merged into a single time-aligned VRS, and the raw color streams are debayered into RGB images.  

We apply a global color correction to the rig videos in order to match the color of the headset RGB cameras.
We also collect and store metadata with each VRS, including calibration data, capture conditions, and video content.

\begin{table}[t]
{\footnotesize
\centering
\begin{tabular}{@{}l@{~~~~}p{6cm}}
\toprule
\textbf{Property} & \textbf{Sample values} \\
\midrule
Lux bins & 51--75, 76--100, 101--200, 201--400, 401--1000, 1001+ \\
Lighting type & natural, artificial, mixed \\
Scene & lab, office, living room, bedroom, rooftop, ... \\
Scene layout & type of furniture present, windows, mirrors, ... \\
Operator action & typing, swiping panels, operating controllers, ... \\
Objects held & controllers, phones, tablet, cup, book, pen, ... \\
Head motion & static, looking up/down/sideways, small/large motion \\
Clothing & short sleeve, long sleeve, t-shirt, suit, hoodie, ... \\
Accessories & watch, rings, colored nails, ... \\
Other person & true, false \\
\bottomrule
\end{tabular}
\caption{Sample dataset properties and axes of diversity. We provide metadata summarizing the properties of each recording.}
\label{tab:dataset_properties}
}
\end{table}

\subsubsection*{Calibration}

We calibrate our rig in a lab using five large planar Calibu targets.  We move the rig through a series of predefined positions, track the location of the calibration markers, and solve for intrinsics and relative extrinsics of all cameras.
During acquisition of our full dataset, which spanned several weeks, we periodically recalibrated our rig to confirm that the calibration parameters remained stable.  When operating in the field, we derive rig poses from the visual inertial odometry (VIO) performed by the headset. 

In addition to this offline calibration procedure, we also perform online calibration to compensate for calibration changes over time.  While camera locations are unlikely to change due to the rig’s stable optical bench, camera orientations can change by 0.1--0.2 degrees due to lens movements, and focal lengths can change with temperature.  This can result in image shifts of 1--3 pixels, which can significantly affect reconstruction accuracy.  Our online calibration refines camera rotations and focal lengths, but keeps other extrinsics and intrinsics fixed.  To perform online calibration, we assume that the calibration remains constant over a full recording.  We detect features and match them across all cameras over the entire recording, but treat each frame separately.  We then optimize camera orientations and focal lengths to jointly minimize reprojection errors of all detected feature points.

We can measure the benefit of online calibration by running a stereo method on different image pairs and measuring the agreement of pairwise depth estimates (see Section~\ref{sec:experiments}).  If the calibration is not accurate, different camera pairs will yield inconsistent depth estimates for the same scene point, which we can measure by computing their median absolute deviation (MAD).  In our experiments, online calibration lowers the median MAD score by 35\%.

\begin{figure}[t]
    \centering
    \includegraphics[width=\linewidth]{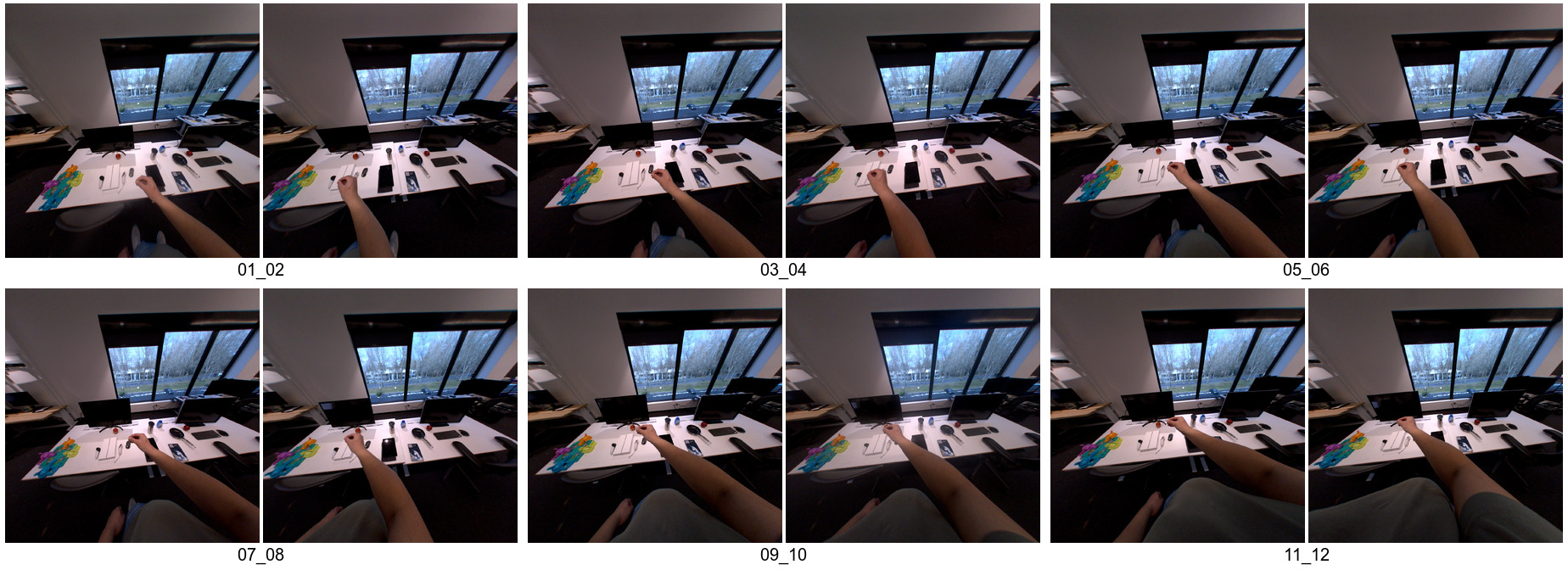}\\[1mm]
    \includegraphics[width=\linewidth]{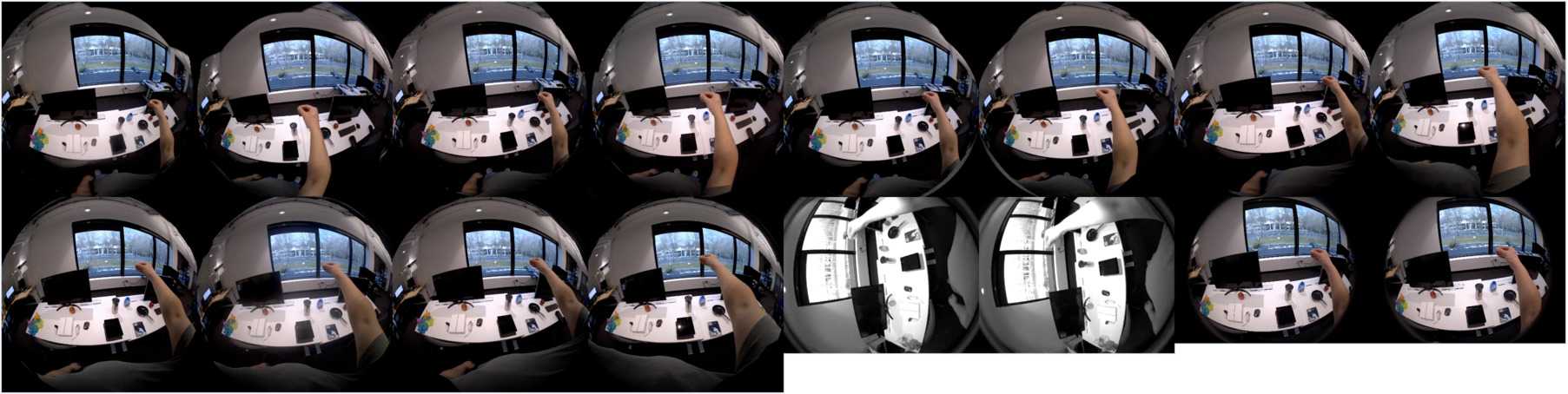}
    \caption{Top: a sample frame from a recording in our research dataset, consisting of 6 rectified stereo pairs.  Bottom: the same frame from the raw VRS with 12+4 fisheye camera streams.}
    \label{fig:sampleframe}
\end{figure}

\subsubsection*{Research dataset}

We provide two versions of our dataset: the ``raw'' version that contains the final merged VRS files with all fisheye camera and sensor data, and a clean ``research dataset,'' that contains only the 12 rig cameras, dewarped into 6 rectified (pinhole) stereo pairs for easy processing.  We omit the headset RGB cameras in the research dataset since they differ from the rig cameras in several aspects: rolling (vs.~global) shutter, resolution, and color profile.  We use a resolution of 1280 × 1280 and a horizontal field of view of 130° when dewarping the fisheye views, which provides a wide field of view while avoiding out-of-bounds regions.  We utilize the refined online calibration when dewarping for maximal accuracy.  We provide each frame as a collection of 12 PNG images together with calibration and pose data.  Fig.~\ref{fig:sampleframe} shows a sample frame in both the raw and the research dataset.  In the next section we show the value of this research dataset for benchmarking of dynamic reconstruction methods.

The average size of a single recording with 450--580 frames from 12 cameras is 19 GB.  The full research dataset with 956 recordings requires 17.5 TB storage and is available at 
\url{https://huggingface.co/datasets/facebook/ego-1k}.
The raw dataset has an average size of 93 GB per VRS for a total of 88 TB;
it is available upon request.
\section{Experiments}
\label{sec:experiments}

We use our dataset, which features challenging dynamic interactions and frequent occlusions, to evaluate existing novel-view synthesis (NVS) methods.  We consider both static NVS methods (run per frame on each set of 12 input images) and dynamic NVS (DNVS) methods, which take the entire dataset (12 views × \mytilde500 frames) as input.  Our experiments demonstrate that the dataset is very challenging and that neither NVS nor DNVS methods can reliably reconstruct the scene in the presence of near-range dynamic hand motions. We also show how using estimated stereo depths as a prior for the 4D reconstruction alleviates the problem of ill-posedness. Below, we first demonstrate how we choose the stereo depth estimation algorithm by measuring consistency among the synchronized frames. Then, we compare different novel-view synthesis algorithms, providing a new baseline for multiview egocentric 4D reconstruction. For all our evaluation experiments, we selected 10\% of our dataset (96 recordings).

\begin{figure}[t]
    \centering
    \includegraphics[width=0.55\linewidth]{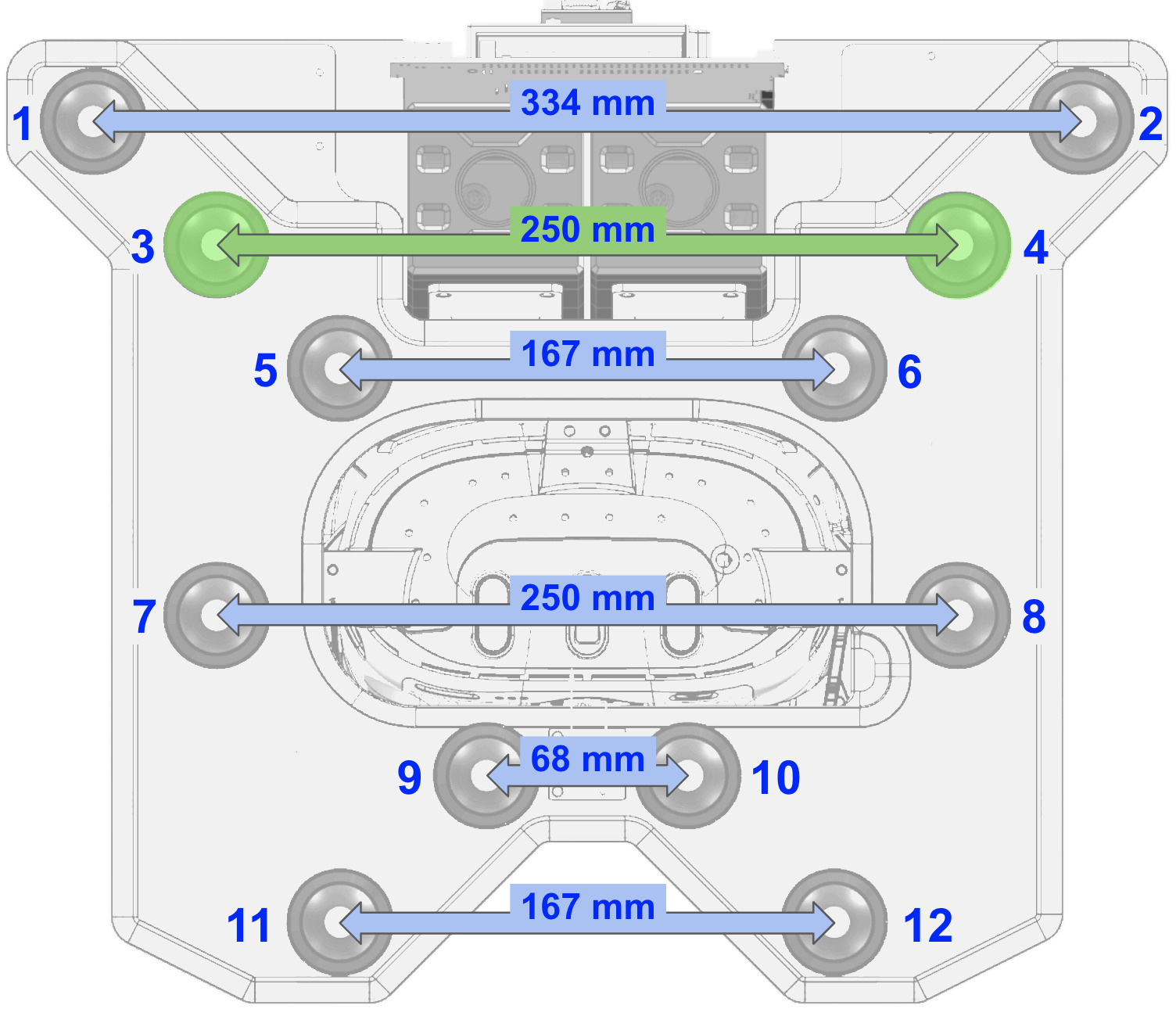}\\
    \caption{Rig stereo pairs and target pair.  The arrows show the 6 rectified stereo pairs; note that the baselines for most pairs are significantly larger than human eye distance (roughly the distance of the headset cameras and pair 9--10).  The target pair (3--4) is shown in green.  We warp the disparity maps of the other 5 pairs to the target pair and evaluate their consistency.}
    \label{fig:stereopairs}
\end{figure}

\subsection{Evaluating pairwise stereo methods}

\begin{table}[t]
\centering
{\small
\begin{tabular}{lccc} 
\toprule
\makecell[l]{Model\\~} &
\makecell{MAD$ \downarrow$ \\ {[mm]}} &
\makecell{MAD \\$<$1mm $\uparrow$} & 
\makecell{SD $\downarrow$\\ {[mm]}} \\
\midrule
Foundation Stereo~\cite{foundationstereo_wen_cvpr25} & \bf 1.6 & \bf 74.0\% &     42.5 \\
Selective-Stereo~\cite{selective_wang_cvpr24}        &     8.0 &  0.0\% &     46.2 \\
BiDAStereo~\cite{bida_jing_eccv24}                   &     2.2 &  3.1\% & \bf  8.3 \\
StereoAnywhere~\cite{anywhere_bartolomei_cvpr25}     &     1.7 & 29.5\% &     10.4 \\
\bottomrule
\end{tabular}
}
\caption{Quantitative evaluation of stereo consistency. MAD: median absolute deviation, SD: standard deviation.}
\label{tab:stereo_consistency}
\end{table}

In order to use stereo depth as a prior for our 4D NVS reconstruction, we need a systematic way to choose the depth estimation algorithm that is most suitable for our use case.  Since we do not have ground-truth depth for evaluation, we instead measure consistency among disparity maps estimated for the same frame. 

\subsubsection*{Experimental setup}

Given the 12 rig camera views in 6 horizontally rectified stereo pairs, we run stereo algorithms on each pair for all frames.  We choose one of the pairs as the target pair, and warp the disparity maps of all other pairs into the target pair (Fig.~\ref{fig:stereopairs}). We then compute median absolute deviation (MAD) and standard deviation (SD) for measuring consistency between the warped pairs.  We warp the disparity maps by projecting pixels to 3D, creating a triangle mesh based on pixel connectivity (discarding triangles that span large depths), and rendering the mesh from the perspective of the target cameras.

\subsubsection*{Results}

We evaluate four recent stereo methods: Foundation Stereo~\cite{foundationstereo_wen_cvpr25}, BiDAStereo~\cite{bida_jing_eccv24}, Selective-Stereo~\cite{selective_wang_cvpr24}, and StereoAnywhere~\cite{anywhere_bartolomei_cvpr25}.  Figure~\ref{fig:qualitative_stereo} visualizes the qualitative results of these stereo methods on our data.  We find that Foundation Stereo yields the most consistent estimates, as measured by per-pixel MAD and SD.  Table~\ref{tab:stereo_consistency} shows a summary of the stereo evaluation on our dataset. We verify that Foundation Stereo has the lowest MAD, while BiDAStereo has the lowest SD. This indicates that BiDAStereo may have less extreme outlier depth values, while in general Foundation Stereo has better consistency. In our next set of experiments, we choose Foundation Stereo based on its superior qualitative results and robust quantitative performance over the other state-of-the-art methods. 

\begin{figure}[t]
    \centering
    \includegraphics[width=0.8\linewidth]{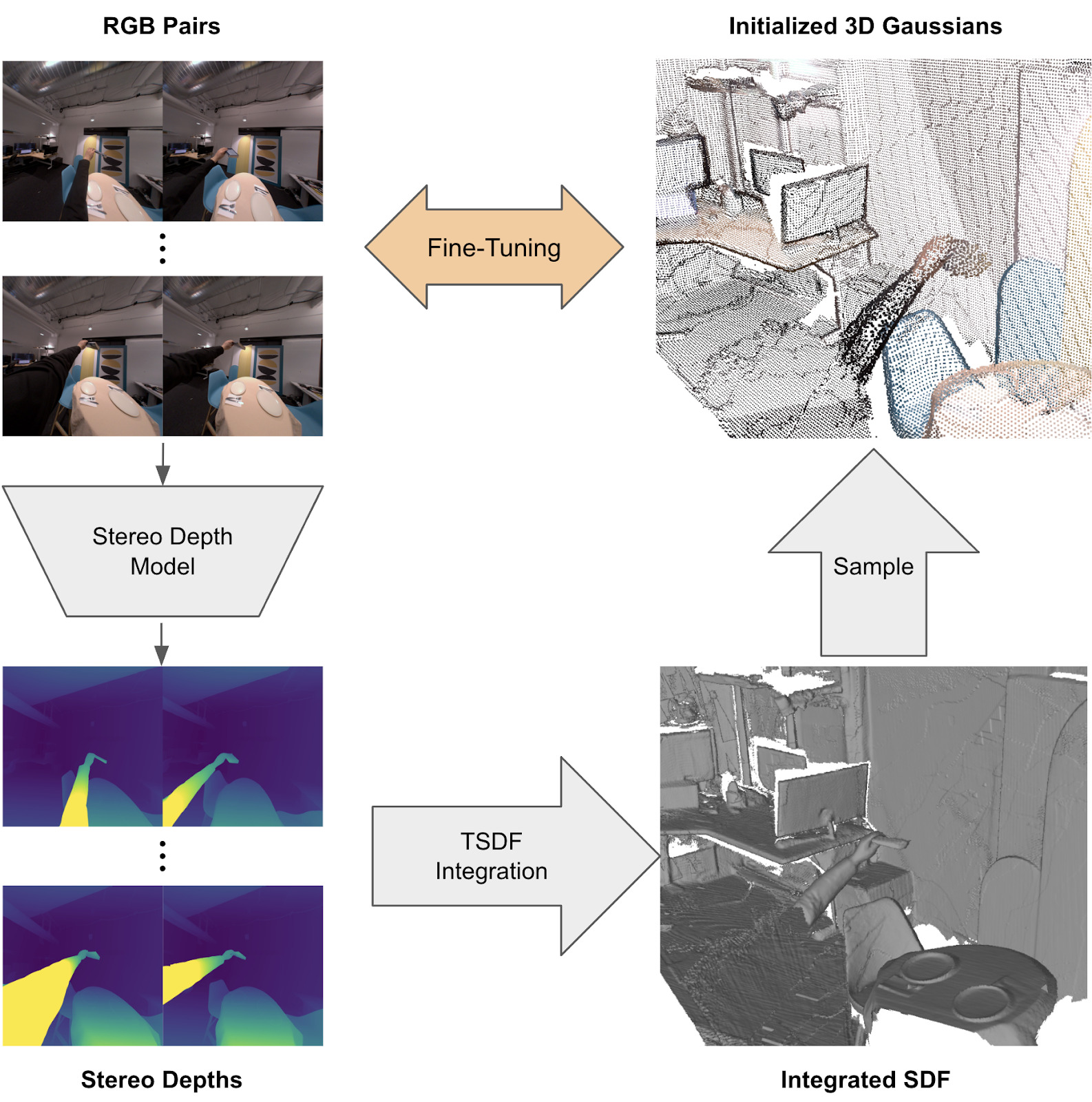}\\
    \caption{
    Stereo-guided 3DGS. We use stereo to compute surfaces, and sample surface points to initialize 3D Gaussians. Then, we fine-tune the 3D Gaussians to minimize photometric loss.}
    \label{fig:stereoguided}
\end{figure}

\newcolumntype{M}[1]{>{\centering\arraybackslash}m{#1}}
\newcolumntype{R}[1]{>{\raggedleft\arraybackslash}m{#1}}
\begin{figure*}[t]
    \centering
    \begin{tabular}{@{}R{6mm}@{\,}M{0.125\textwidth}@{}R{1cm}@{~}M{0.125\textwidth}@{}M{0.125\textwidth}@{}M{0.125\textwidth}@{}M{0.125\textwidth}@{}M{0.125\textwidth}@{}M{0.125\textwidth}}
    3&
        \includegraphics[width=0.125\textwidth]{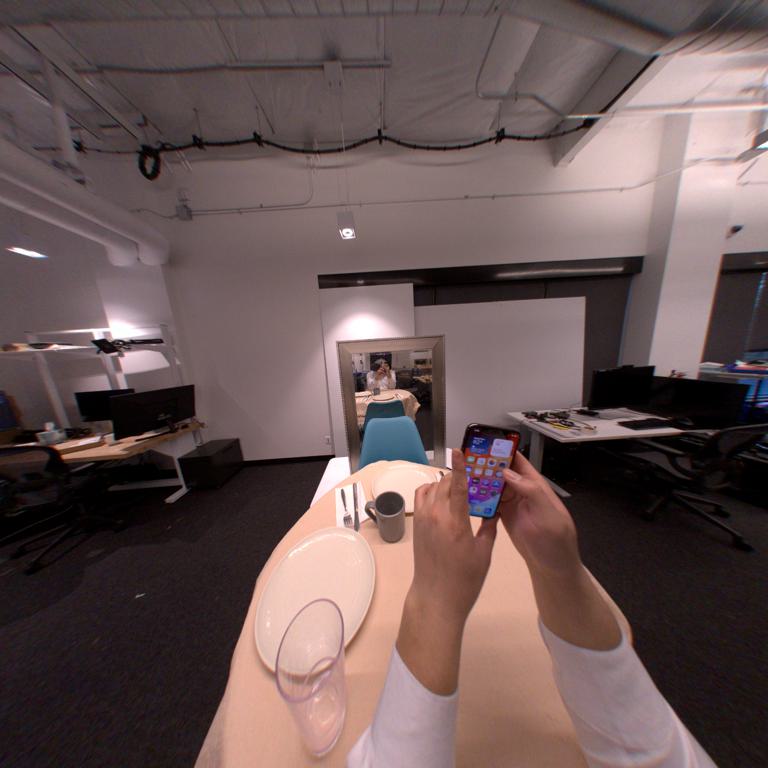} &
        FS & 
        \includegraphics[width=0.125\textwidth]{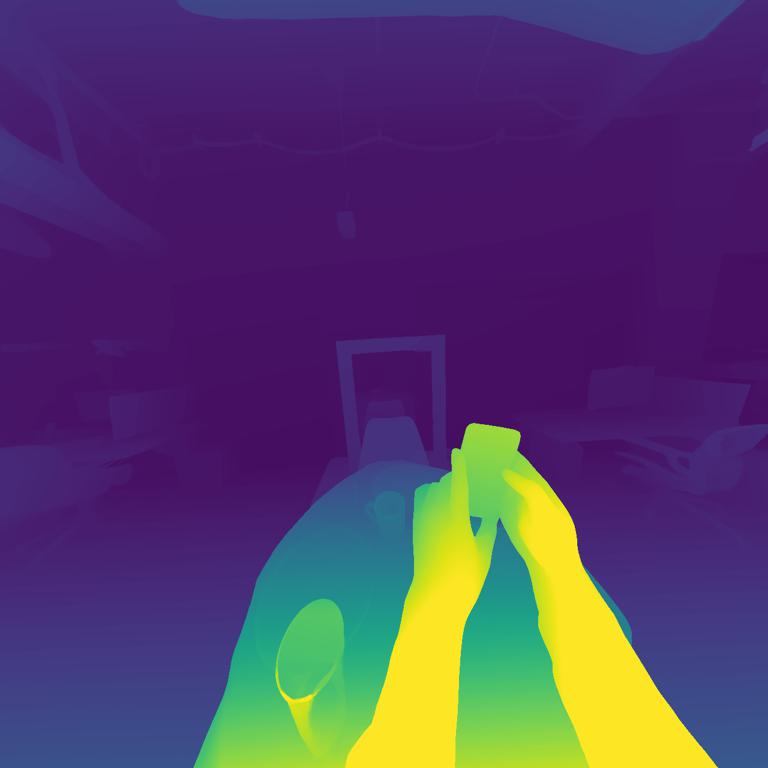} &
        \includegraphics[width=0.125\textwidth]{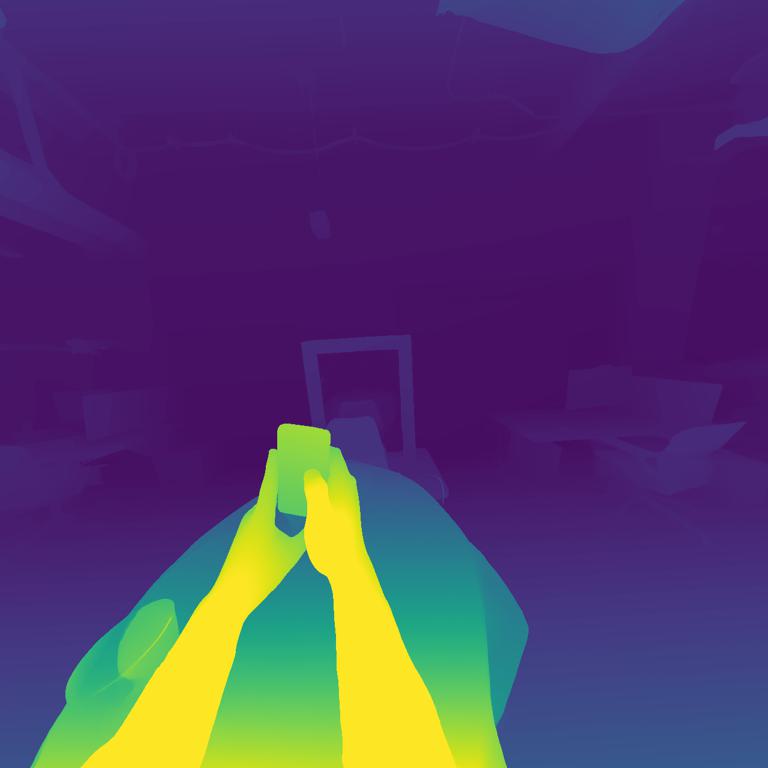} &
        \includegraphics[width=0.125\textwidth]{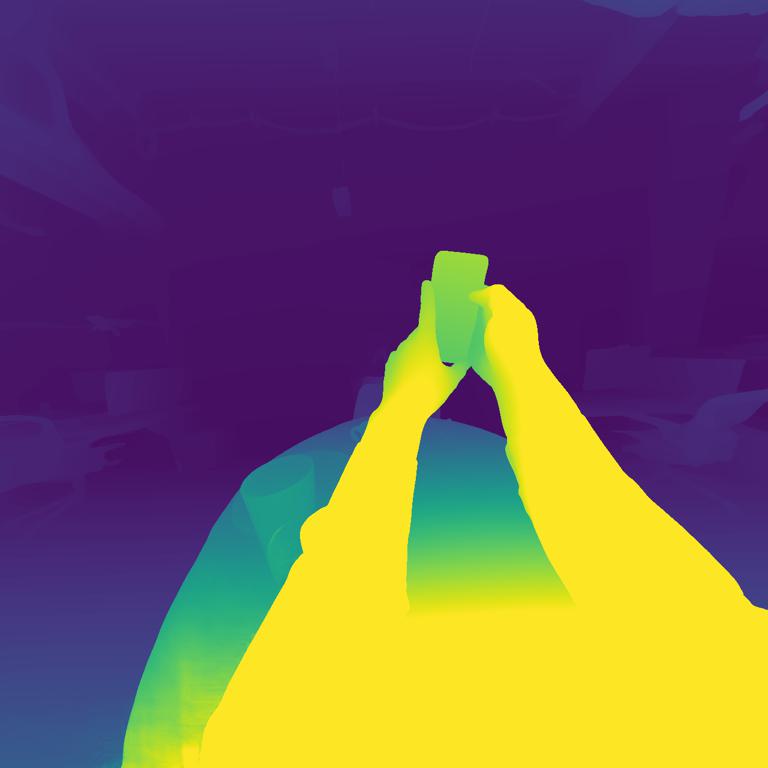} &
        \includegraphics[width=0.125\textwidth]{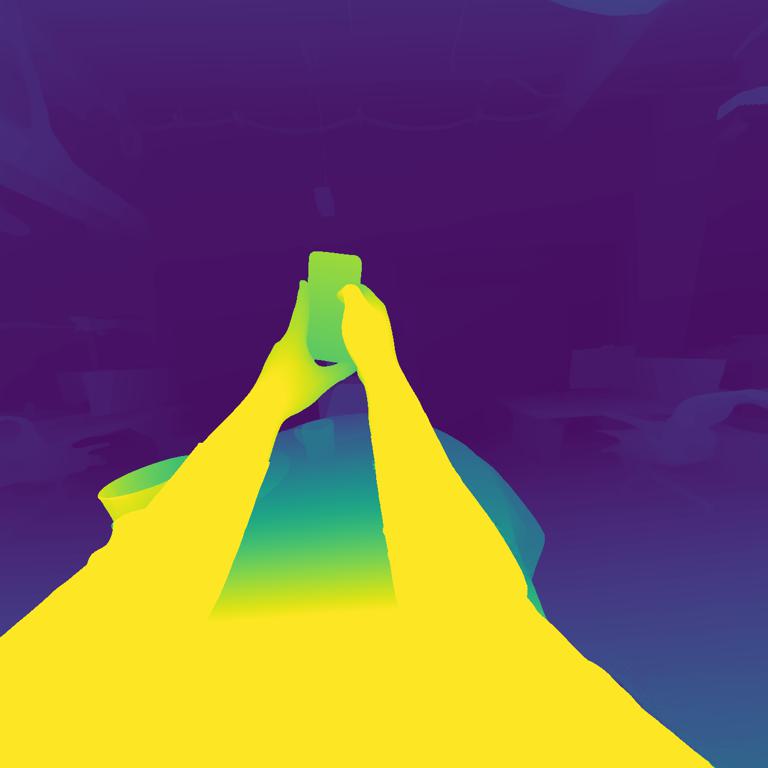} &
        \includegraphics[width=0.125\textwidth]{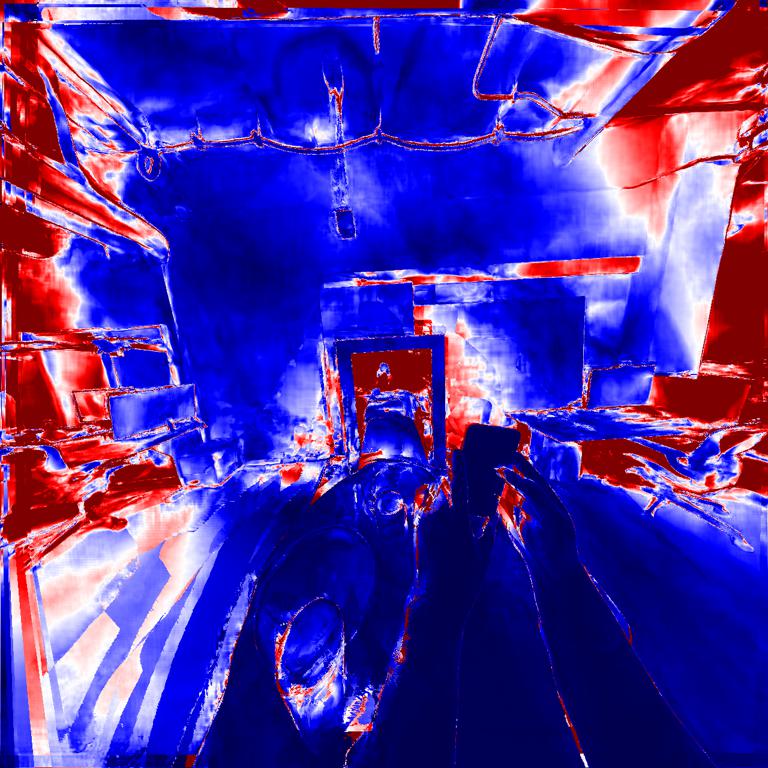} &
        \includegraphics[width=0.125\textwidth]{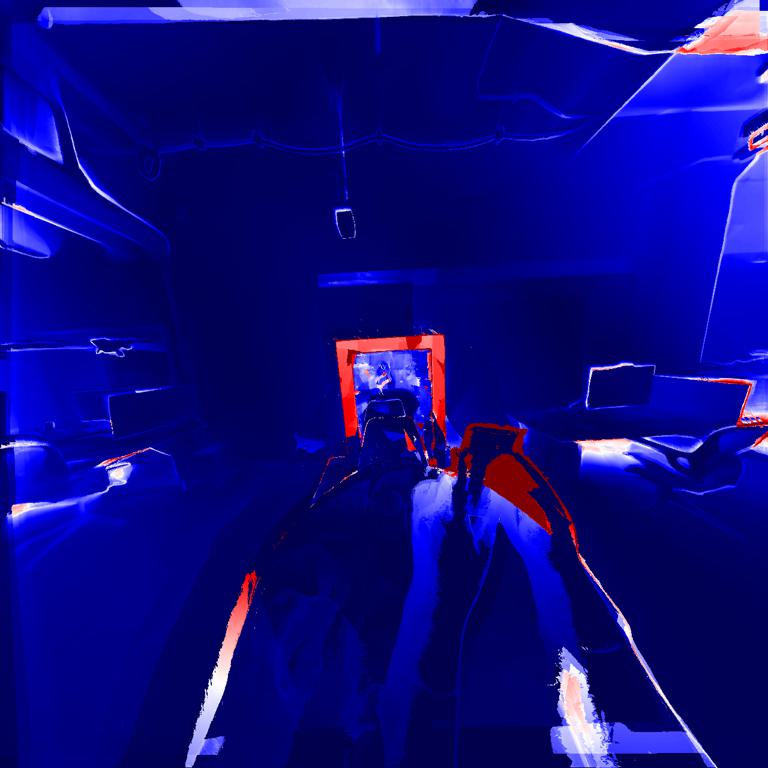} \\[-2pt]
        4 & 
        \includegraphics[width=0.125\textwidth]{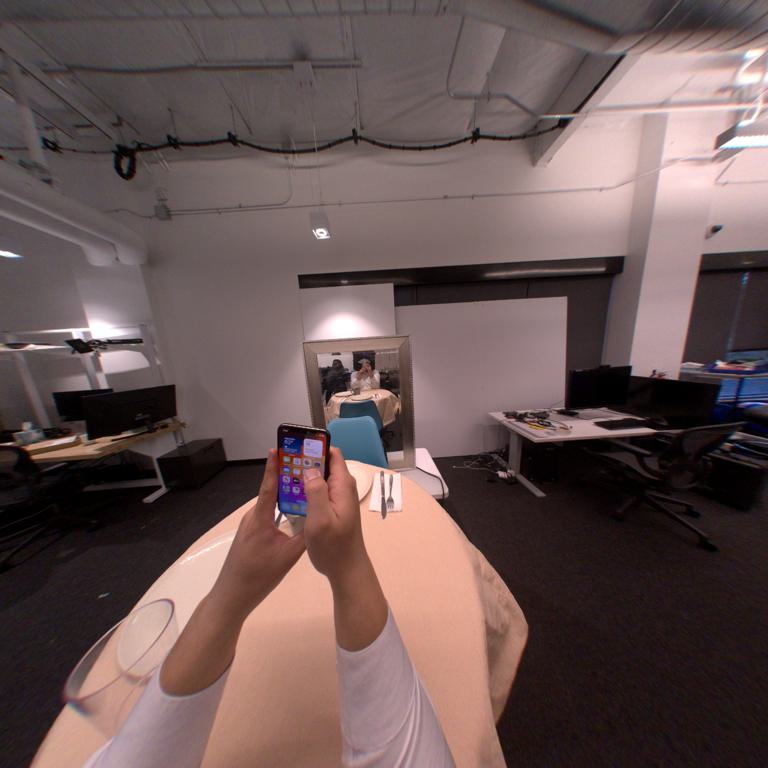} &
        BiDA & 
        \includegraphics[width=0.125\textwidth]{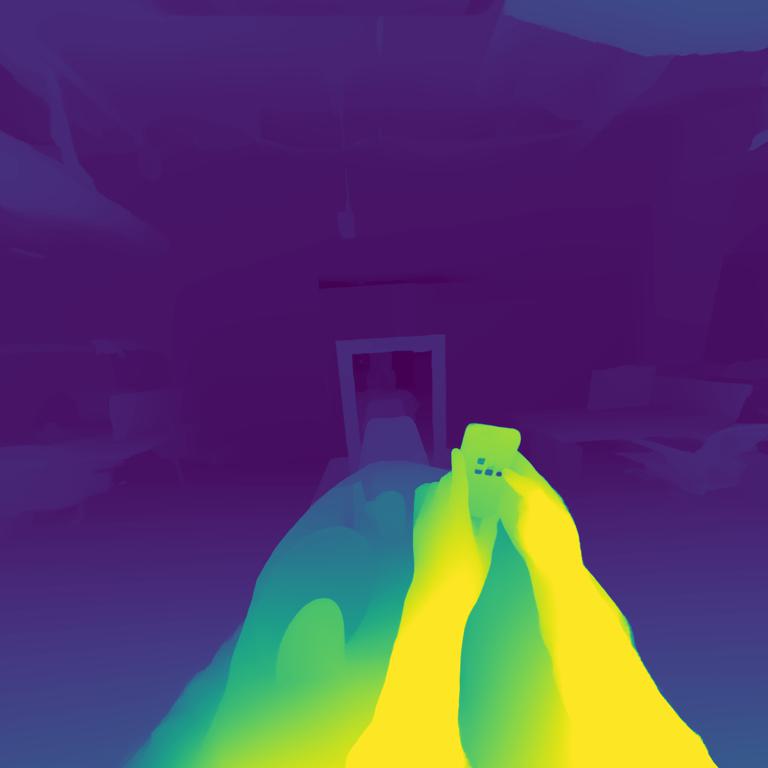} &
        \includegraphics[width=0.125\textwidth]{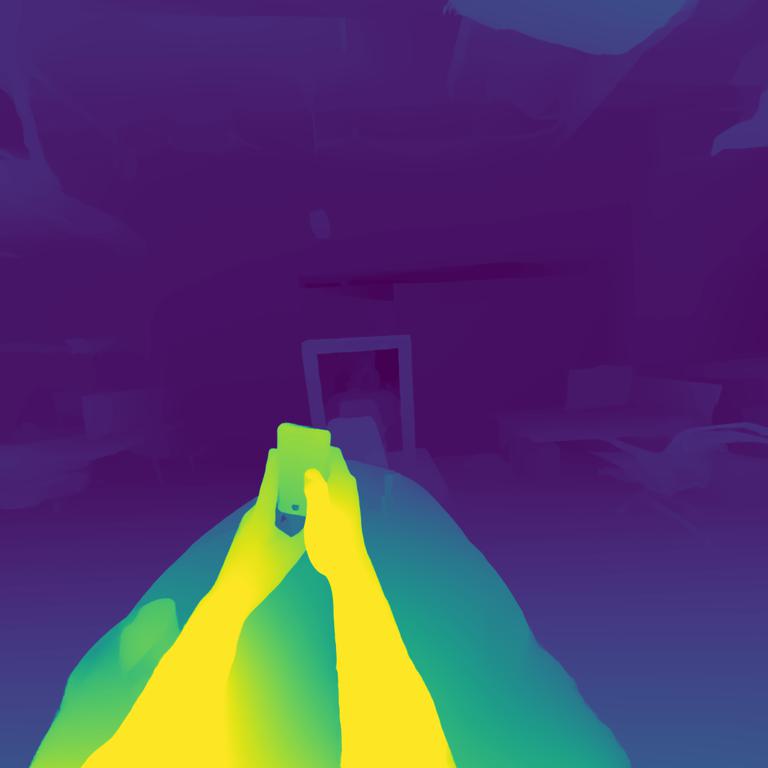} &
        \includegraphics[width=0.125\textwidth]{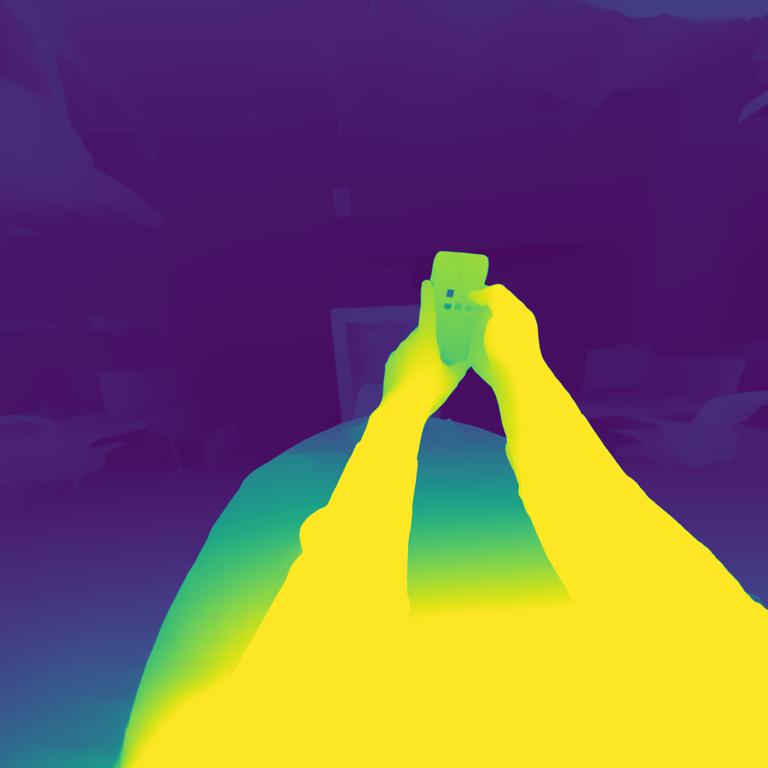} &
        \includegraphics[width=0.125\textwidth]{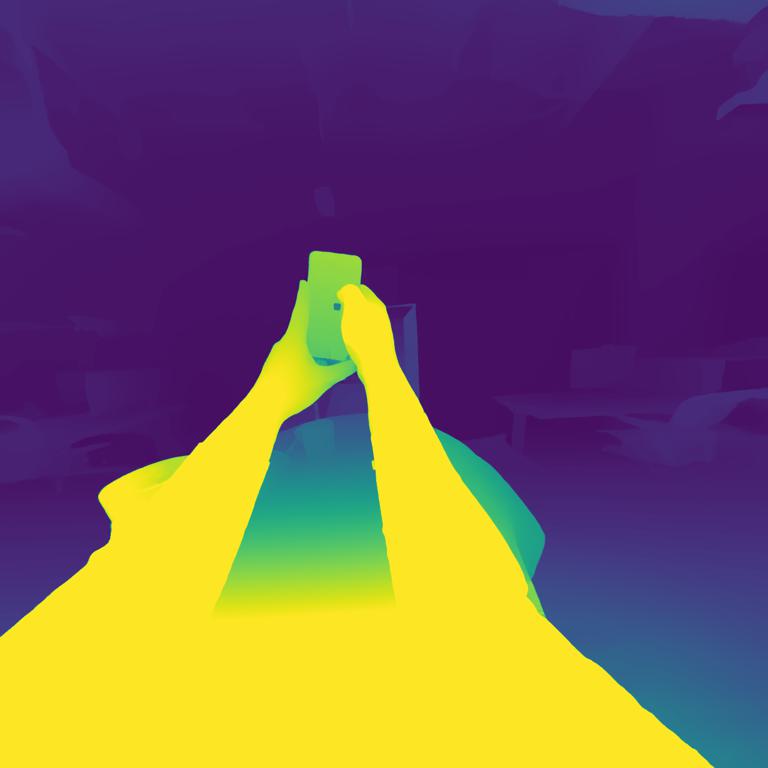} &
        \includegraphics[width=0.125\textwidth]{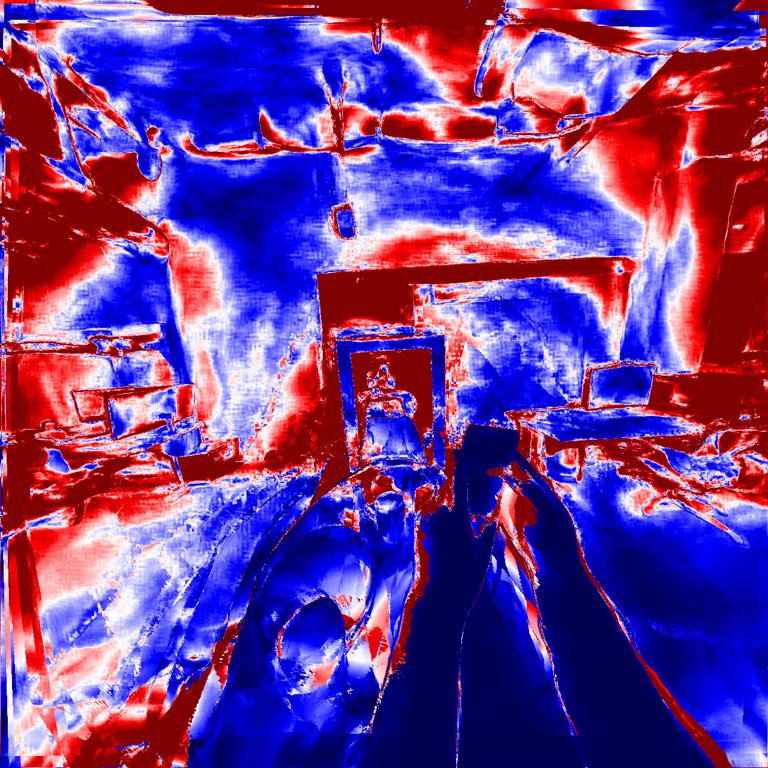} &
        \includegraphics[width=0.125\textwidth]{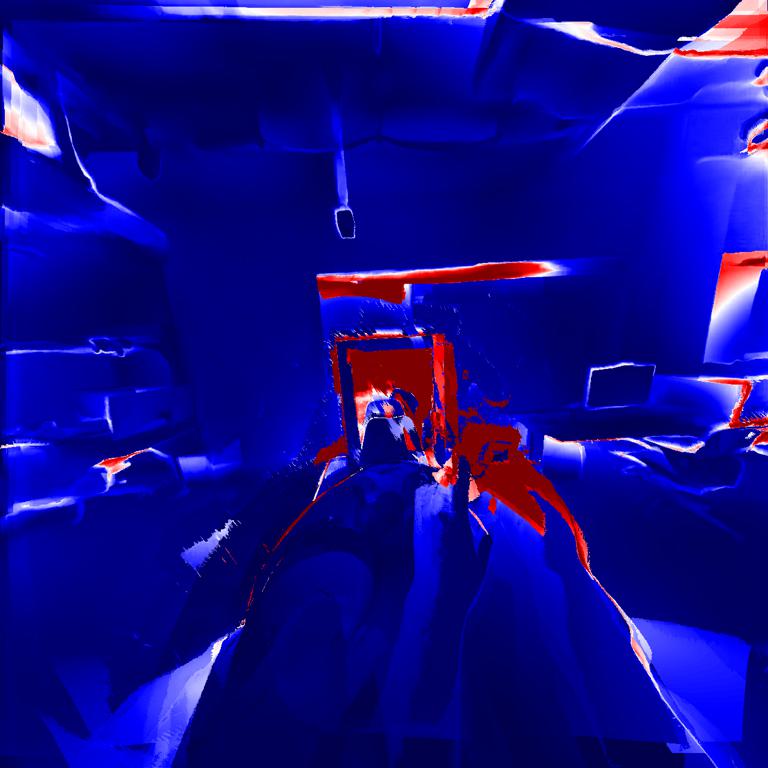} \\[-2pt]
        11 &
        \includegraphics[width=0.125\textwidth]{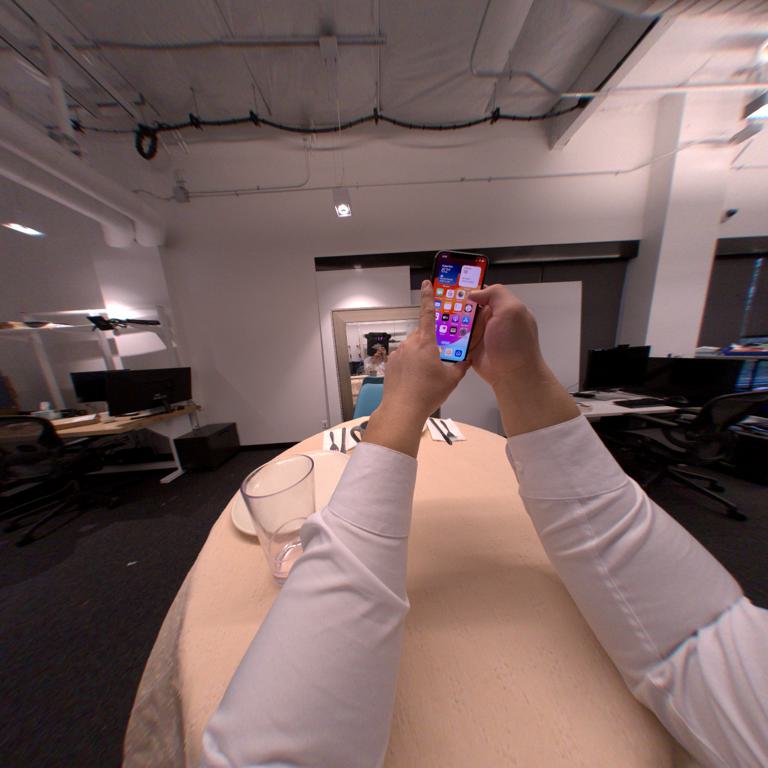} &
        SelS & 
        \includegraphics[width=0.125\textwidth]{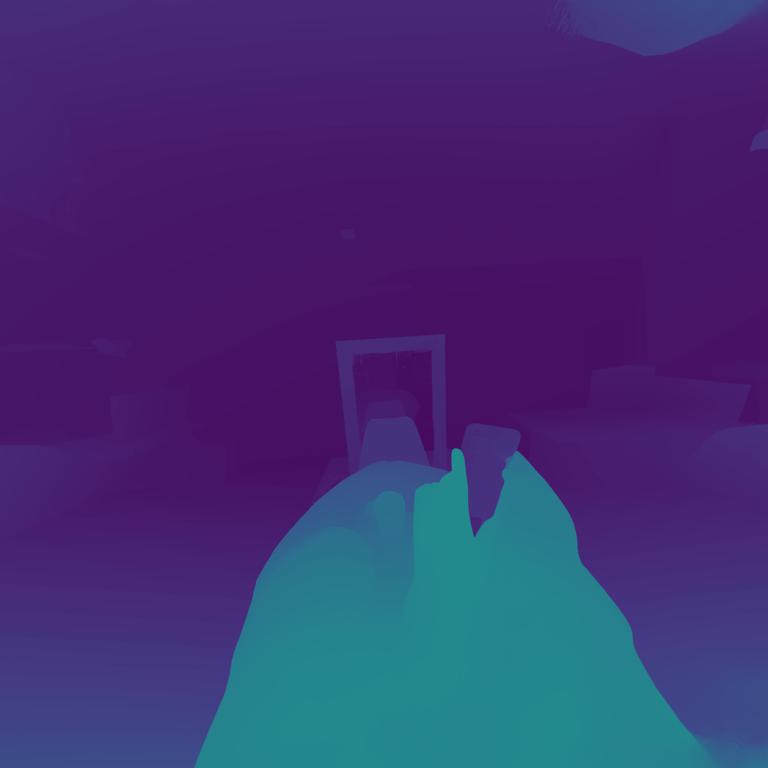} &
        \includegraphics[width=0.125\textwidth]{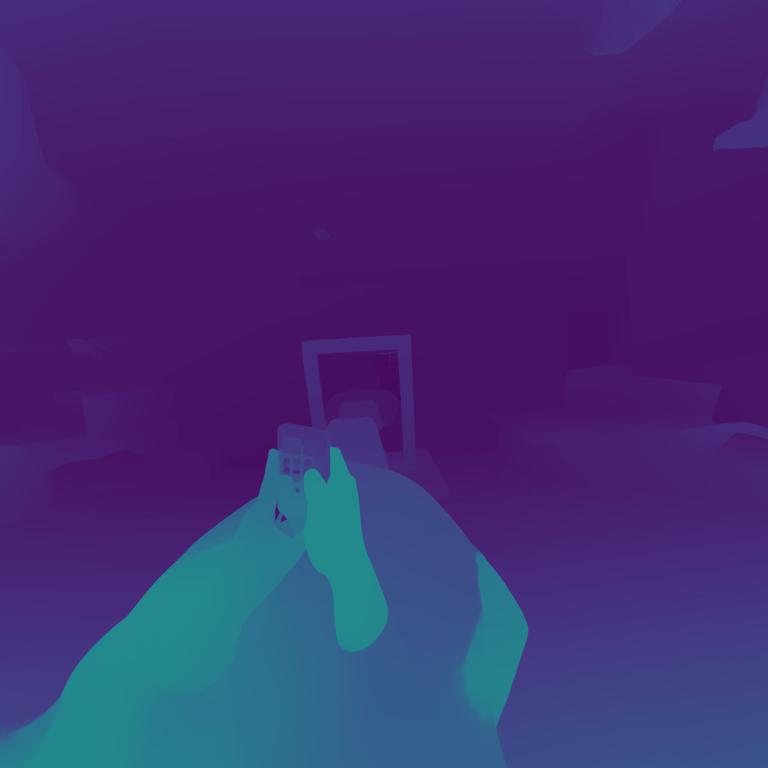} &
        \includegraphics[width=0.125\textwidth]{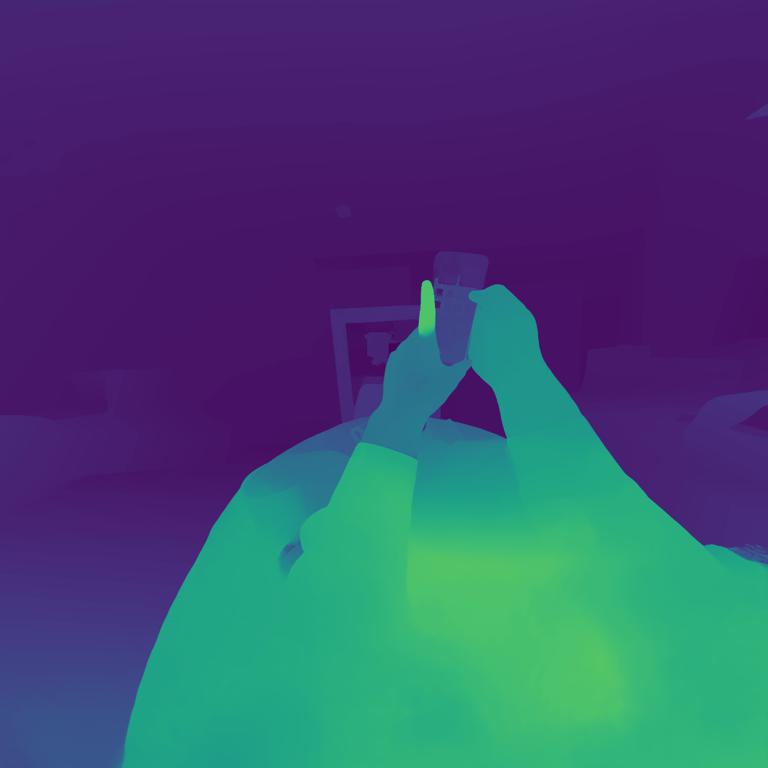} &
        \includegraphics[width=0.125\textwidth]{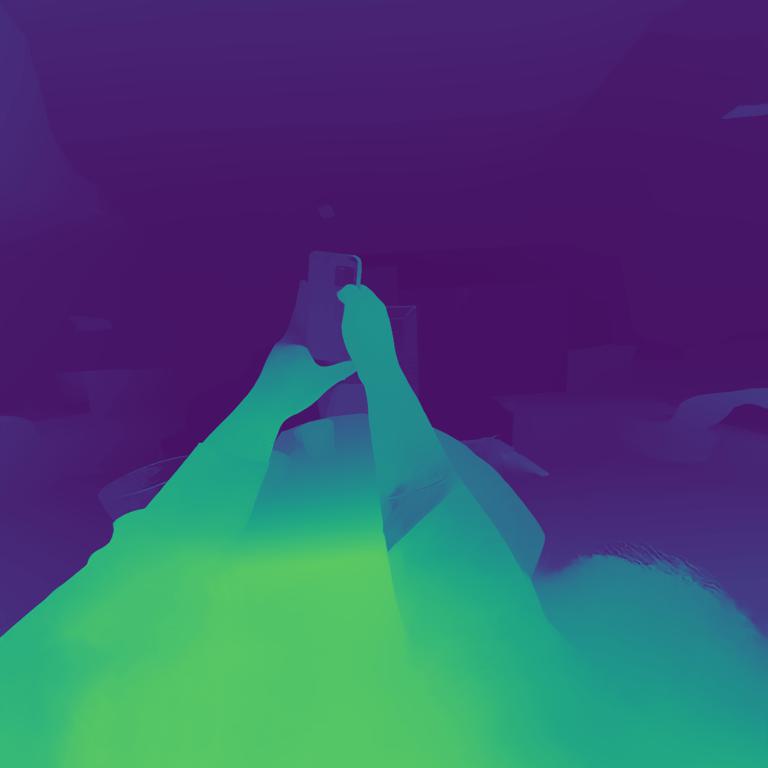} &
        \includegraphics[width=0.125\textwidth]{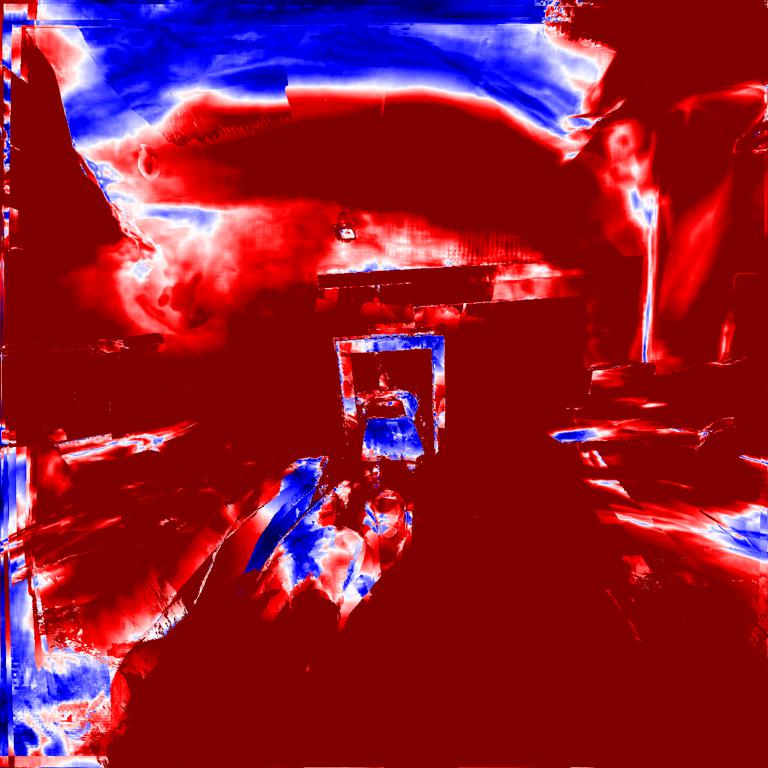} &
        \includegraphics[width=0.125\textwidth]{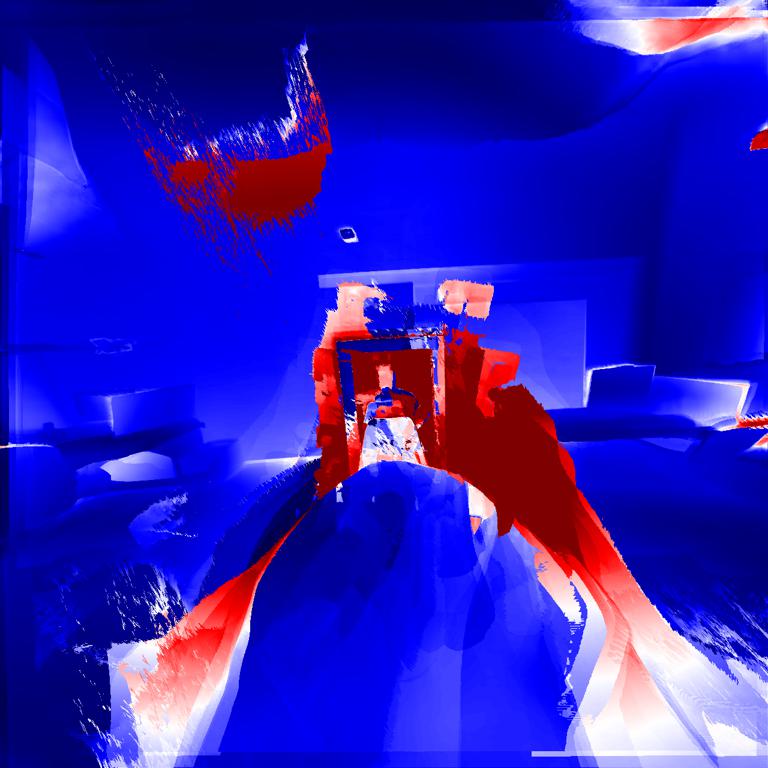} \\[-2pt]
        12 &
        \includegraphics[width=0.125\textwidth]{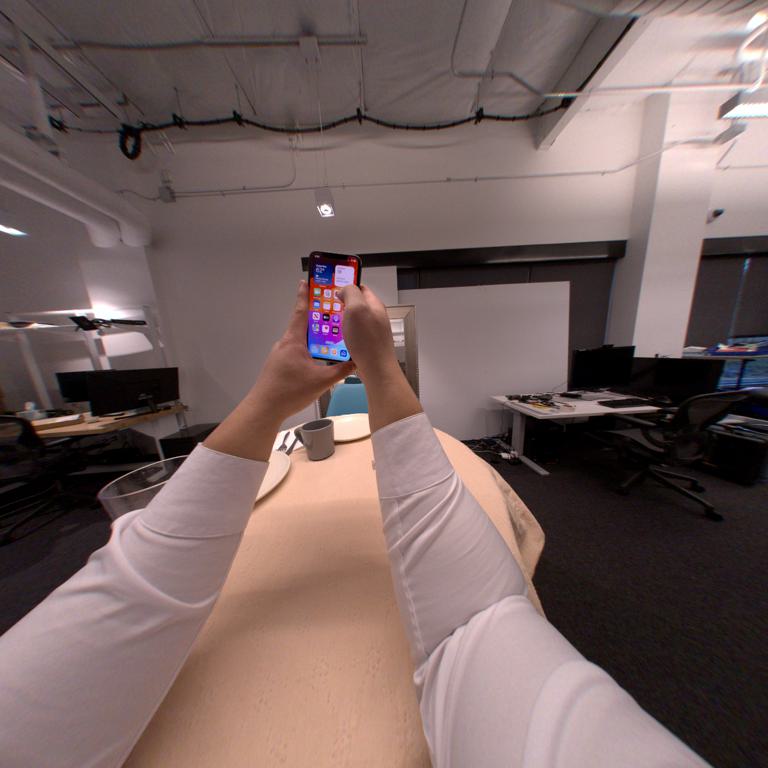} &
        SA & 
        \includegraphics[width=0.125\textwidth]{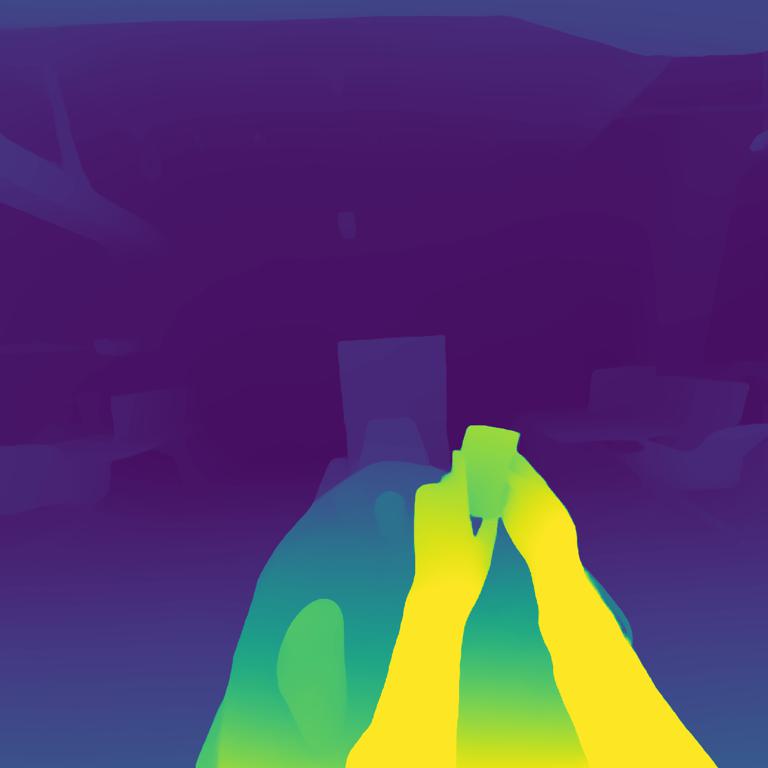} &
        \includegraphics[width=0.125\textwidth]{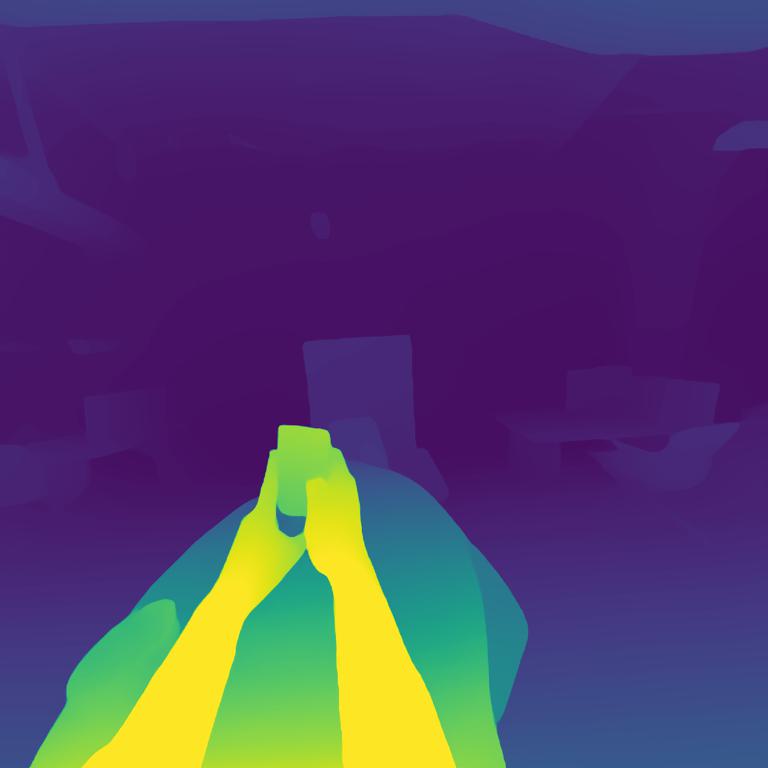} &
        \includegraphics[width=0.125\textwidth]{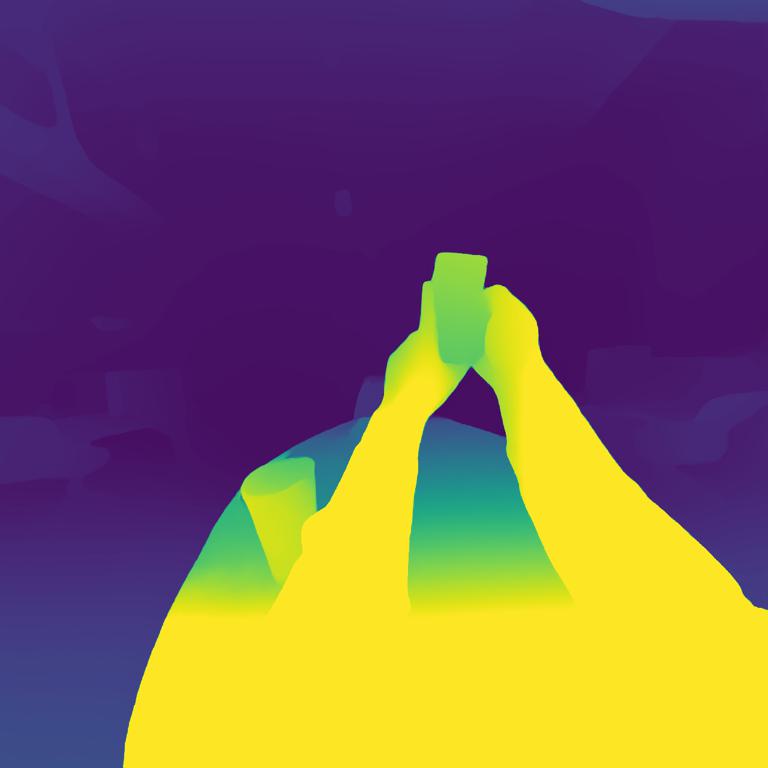} &
        \includegraphics[width=0.125\textwidth]{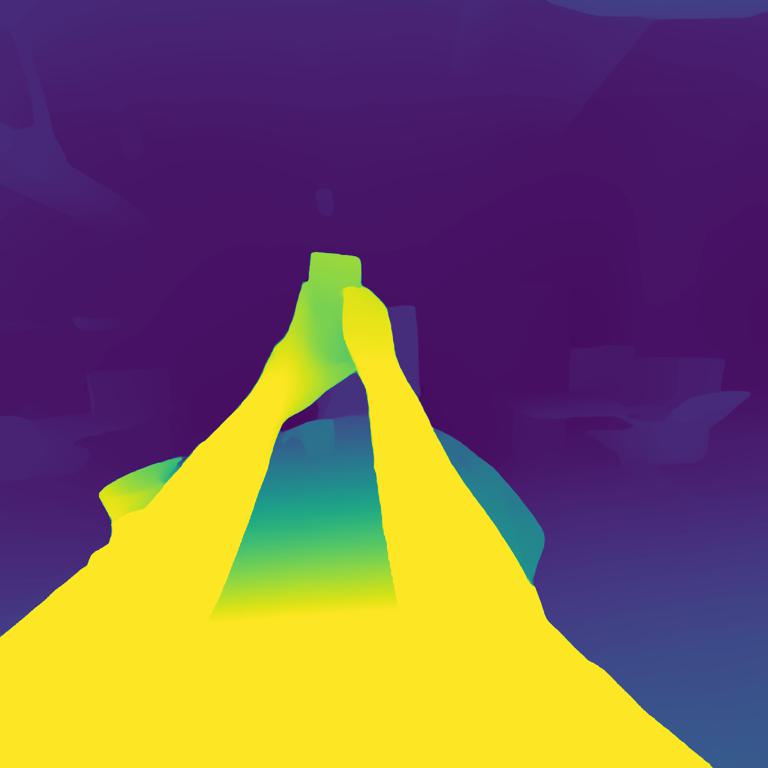} &
        \includegraphics[width=0.125\textwidth]{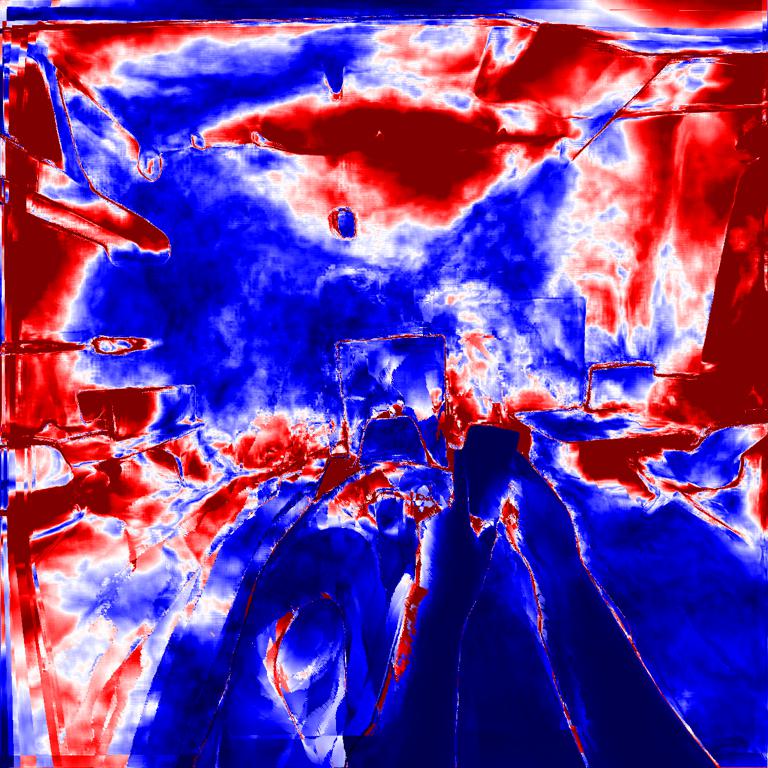} &
        \includegraphics[width=0.125\textwidth]{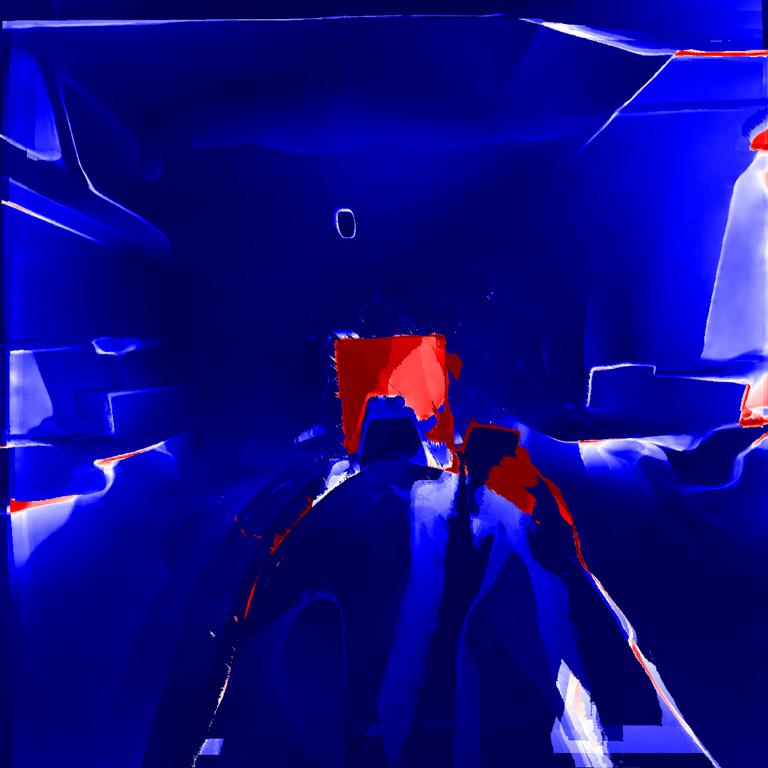} \\
        & RGB & & \multicolumn{4}{c}{Disparities (cams 3-4, 11-12)} & MAD & SD
    \end{tabular}
    \caption{
        Qualitative stereo results. Left: target and bottom-most stereo pairs 3--4, 11--12. Middle/right: stereo results for
        Foundation Stereo (FS)~\cite{foundationstereo_wen_cvpr25}, 
        BiDAStereo (BiDA)~\cite{bida_jing_eccv24}, 
        Selective-Stereo (SelS)~\cite{selective_wang_cvpr24}, and
        StereoAnywhere (SA)~\cite{anywhere_bartolomei_cvpr25}: Disparities; median absolute deviation (MAD) and standard deviation (SD) of all pairs warped to the target camera. We clamp MAD, SD to 4cm, 50cm. 
    }
    \label{fig:qualitative_stereo}
\end{figure*}

\subsection{Stereo initialization for 4D reconstruction}
\label{sec:geom_priors}

Given stereo disparity estimates for each frame, we use them as a geometric prior to improve the 4D reconstruction process.  Unlike existing methods that optimize Gaussian splats with loss-based supervision~\cite{stereogs_safadoust_bmvc24} or use depth as a prior in an end-to-end feed-forward model (e.g.,~\cite{depthsplat_xu_cvpr25}), we find it sufficient to initialize Gaussian splats using precise fused stereo geometry and then fine-tune for a small number of iterations by minimizing photometric loss.  Figure~\ref{fig:stereoguided} provides a high-level overview of our method. Given RGB input pairs, we run Foundation Stereo in both directions (left-to-right and right-to-left).  Next, we fuse all stereo depth maps using TSDF integration~\cite{curless_tsdf_fusion}.  We then sample surface points from the integrated TSDF to obtain a surface normal and color for each sampled point.  Because high-quality fused stereo geometry already translates into high-fidelity 3D Gaussians, we simply fine-tune the Gaussians for a small number of steps, minimizing photometric reconstruction loss~\cite{gs3d_kerbl_siggraph23} with default $\lambda=0.1$: 
\begin{equation}
    \mathcal{L} = (1 - \lambda) \mathcal{L}_1 + \lambda \mathcal{L}_{\text{D-SSIM}}.
\end{equation}
We apply this static-scene optimization independently to every frame, resulting in a dense 4D reconstruction.

\subsection{Evaluating egocentric 4D NVS reconstruction}

To evaluate 4D reconstruction, we fit NVS and DNVS models to our dataset. We start by fitting the state-of-the-art static scene NVS algorithm, 3DGS~\cite{gs3d_kerbl_siggraph23}, for each frame independently. Then, we pick one model based on dynamic radiance fields (K-Planes~\cite{kplanes_fridovich_cvpr23}) and one dynamic 3DGS variant (Spacetime Gaussians~\cite{spacetime_li_cvpr24}) as additional baselines for comparison. Finally, we fit the 3DGS model with additional guidance from stereo geometry to demonstrate that geometry guidance is essential for achieving high-quality results on our dataset.

\subsubsection*{Experimental setup}

Similar to the stereo consistency measurement, we split the cameras into two groups, for training and test.  As before, we use the target pair 3--4 (green cameras in Fig.~\ref{fig:stereopairs}) as the test views, and use the remaining 10 views as training views. We evaluate 4D NVS reconstruction using standard image reconstruction measures (PSNR, SSIM and LPIPS)~\cite{nerf_mildenhall_eccv20} for each frame and camera in the target pair. 

\begin{figure*}[t]
    \centering
    \resizebox{0.922\textwidth}{!}{
    
    \begin{tabular}{c@{}c@{}c@{}c@{}c}
        \includegraphics[width=0.2\textwidth]{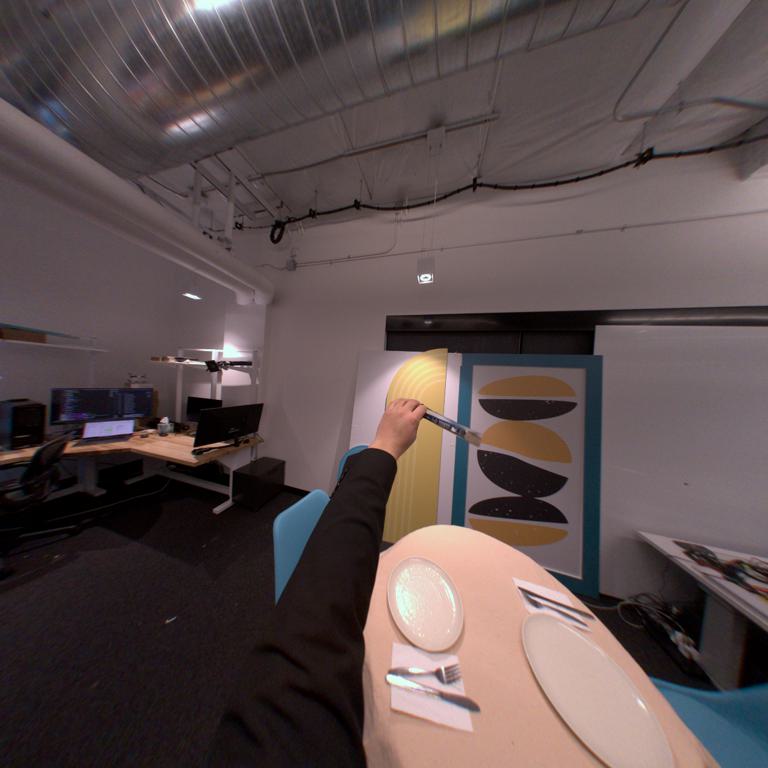} &
        \includegraphics[width=0.2\textwidth]{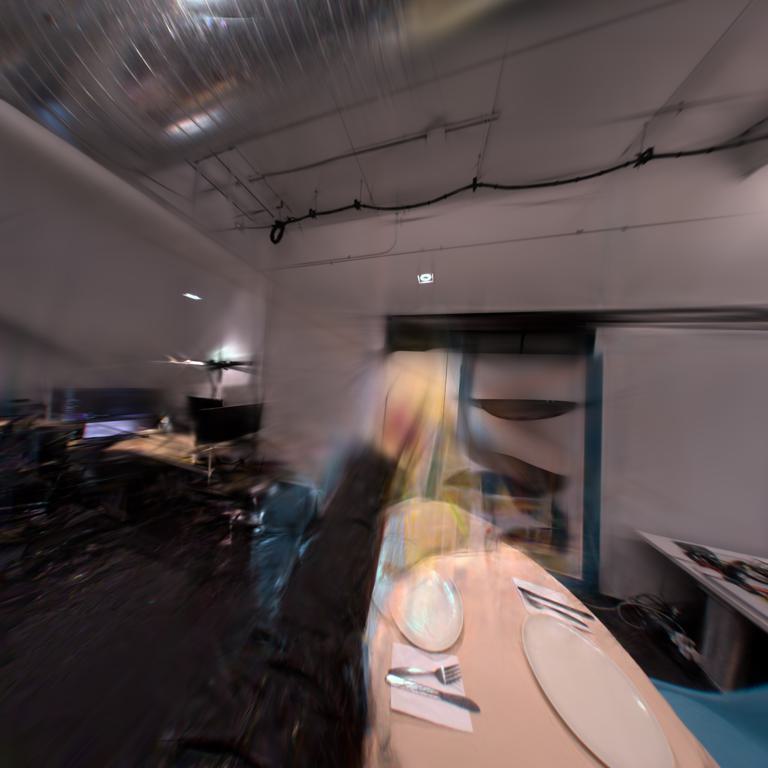} &
        \includegraphics[width=0.2\textwidth]{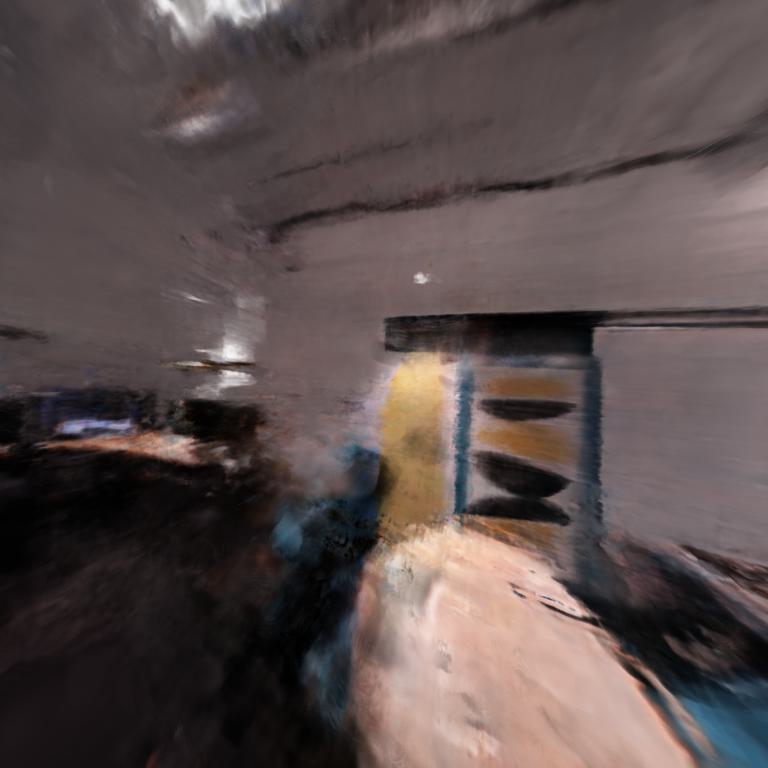} &
        \includegraphics[width=0.2\textwidth]{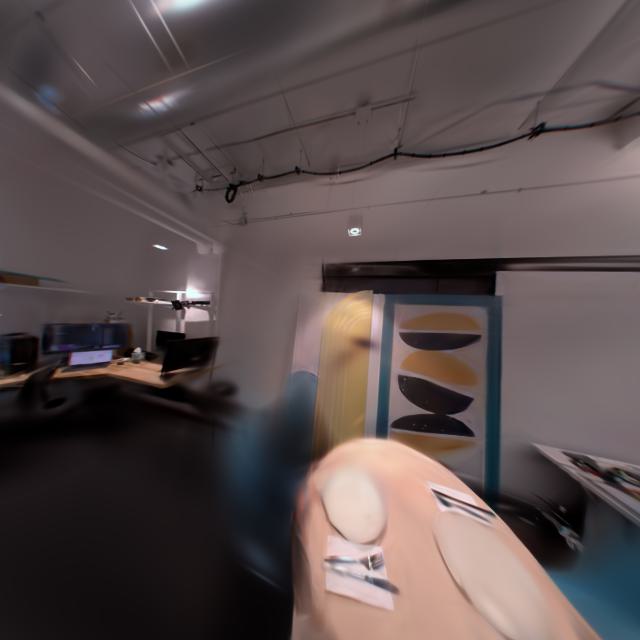} &
        \includegraphics[width=0.2\textwidth]{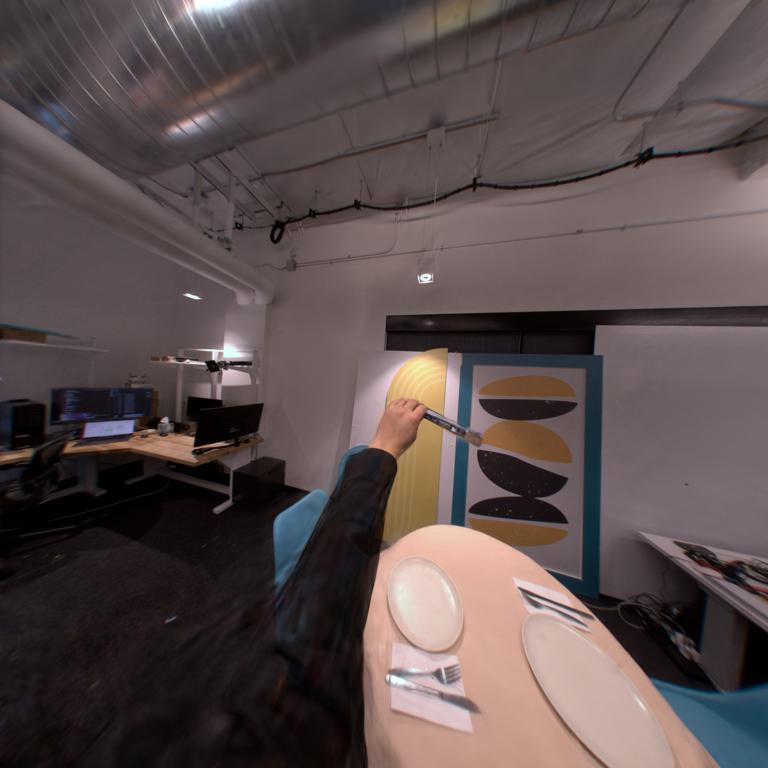}\\[-5pt]
        \includegraphics[width=0.2\textwidth]{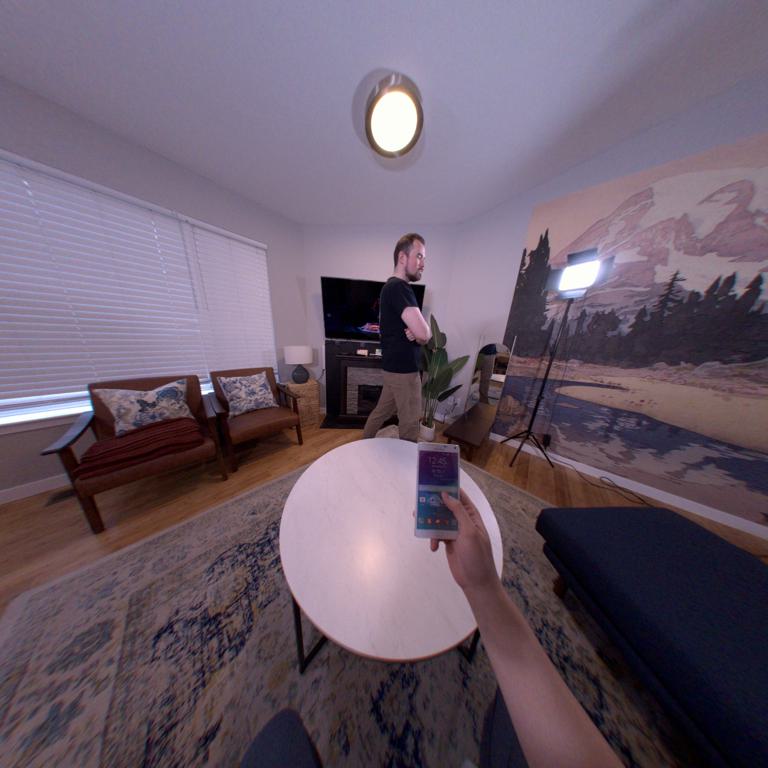} &
        \includegraphics[width=0.2\textwidth]{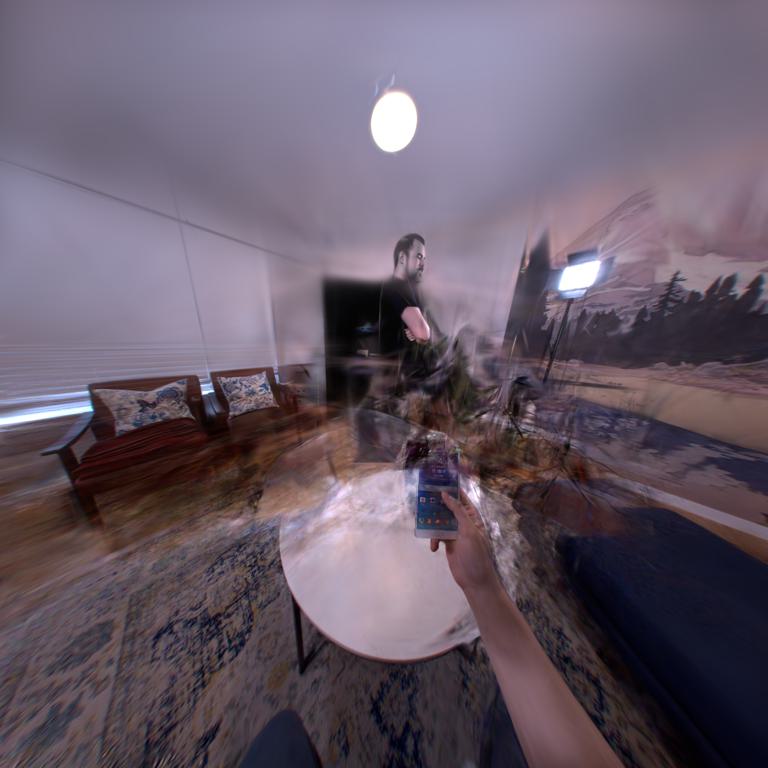} &
        \includegraphics[width=0.2\textwidth]{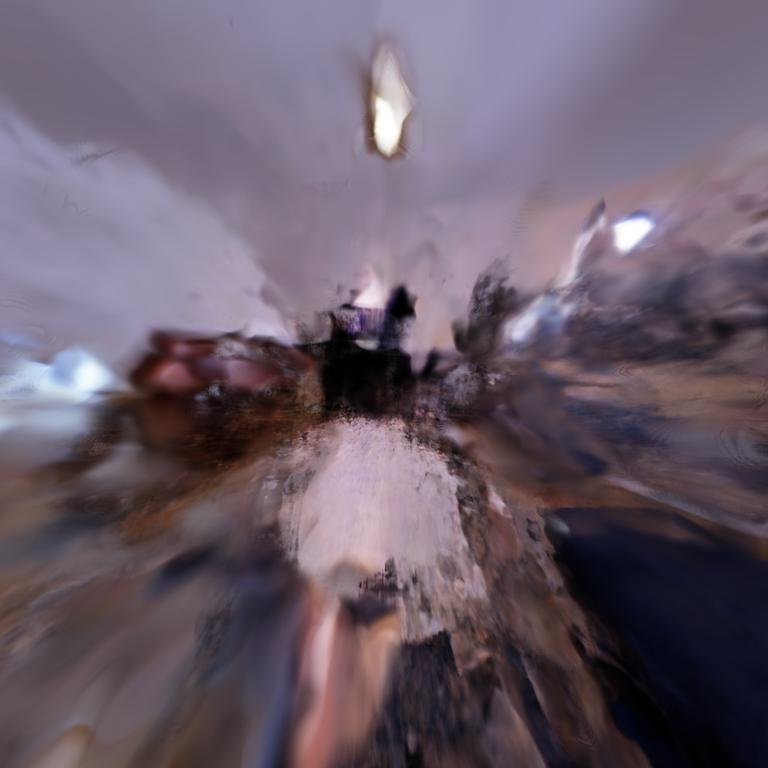} &
        \includegraphics[width=0.2\textwidth]{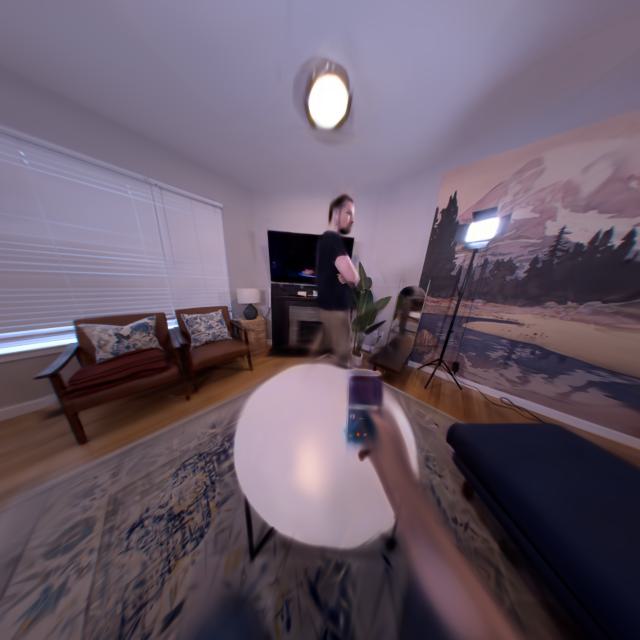} &
        \includegraphics[width=0.2\textwidth]{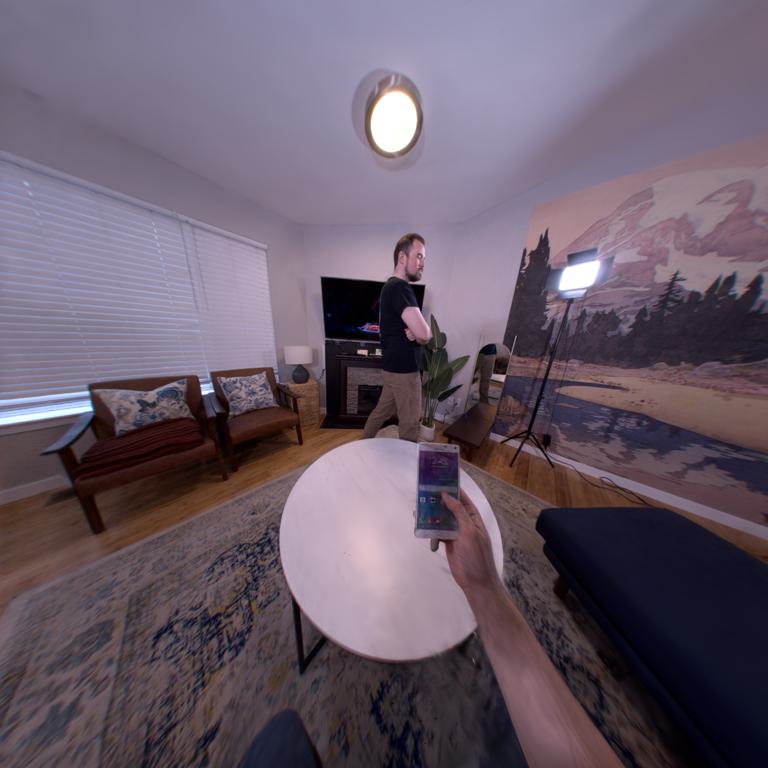}\\[-5pt]
        \includegraphics[width=0.2\textwidth]{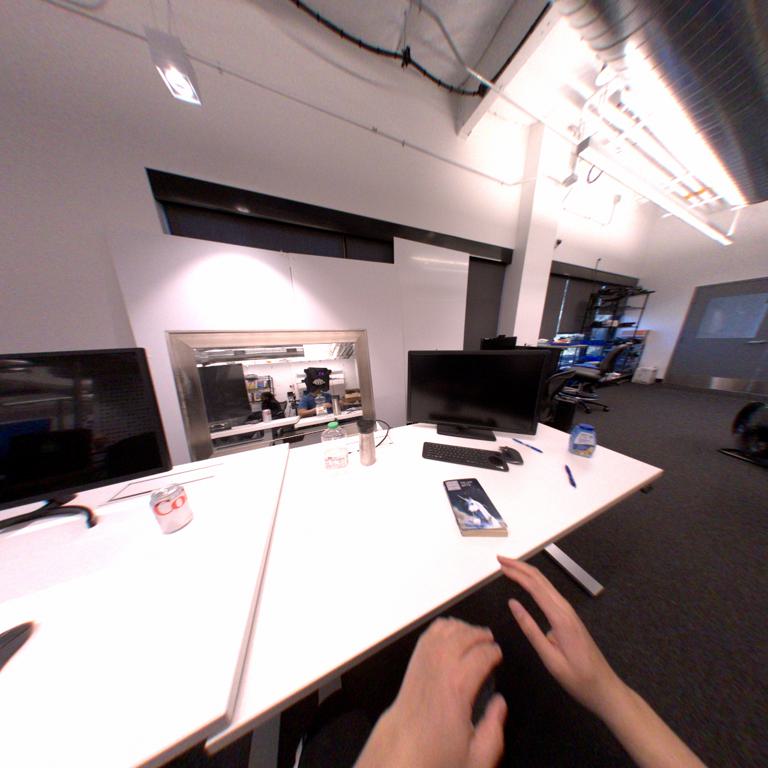} &
        \includegraphics[width=0.2\textwidth]{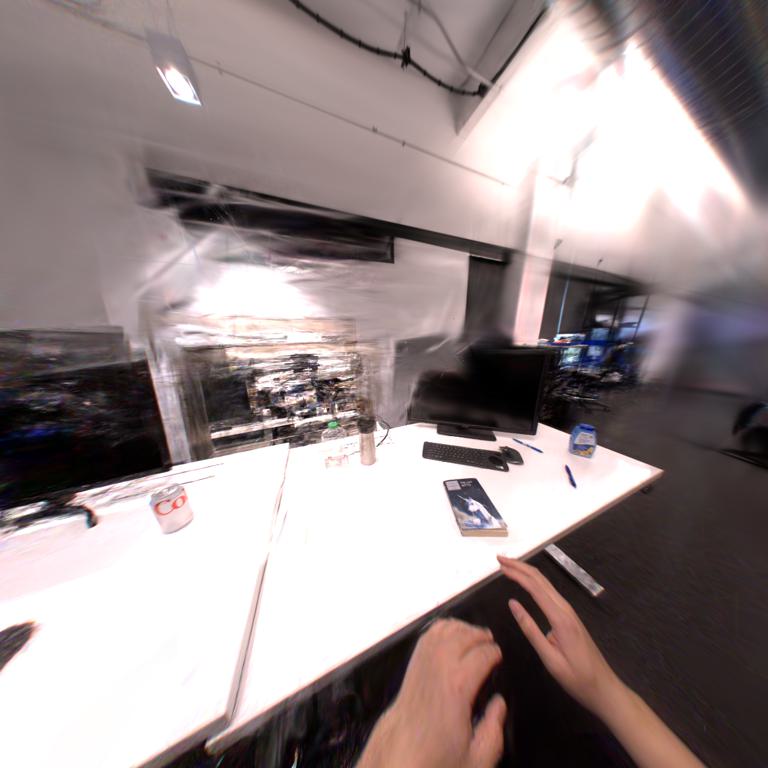} &
        \includegraphics[width=0.2\textwidth]{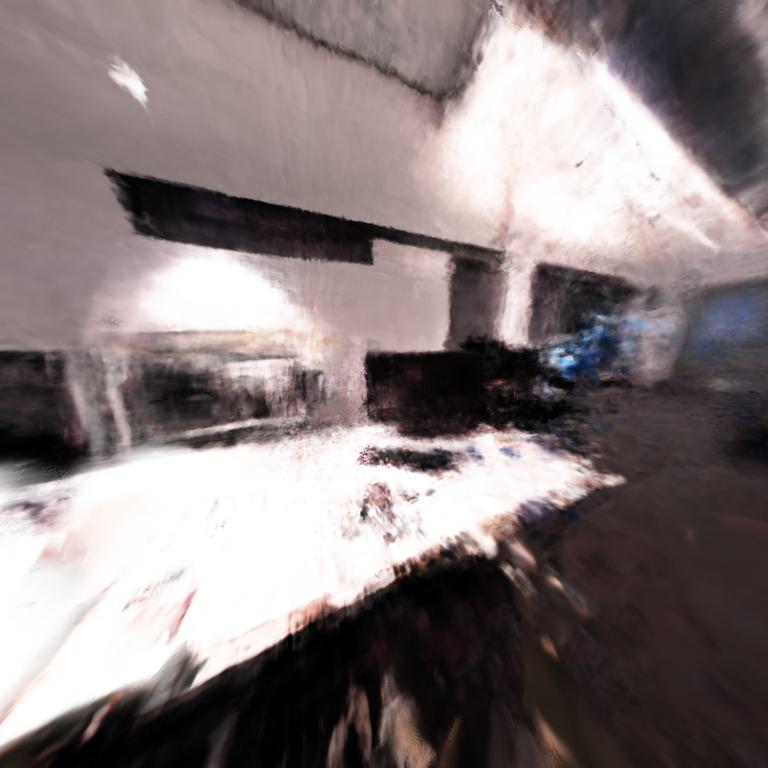} &
        \includegraphics[width=0.2\textwidth]{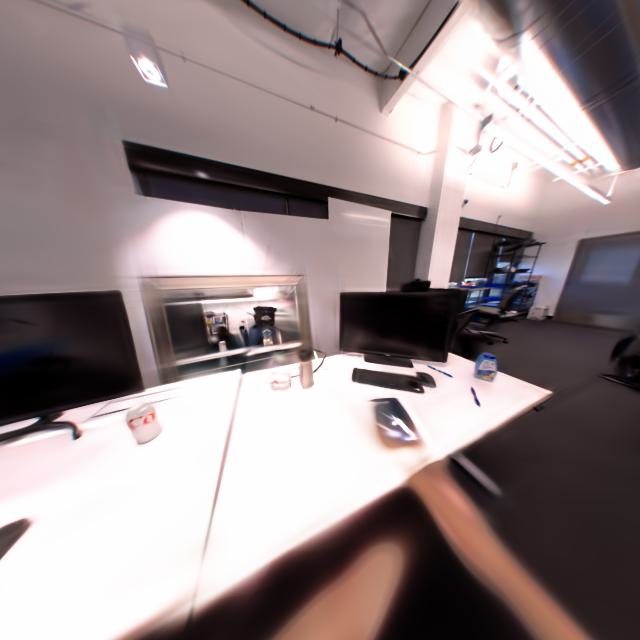} &
        \includegraphics[width=0.2\textwidth]{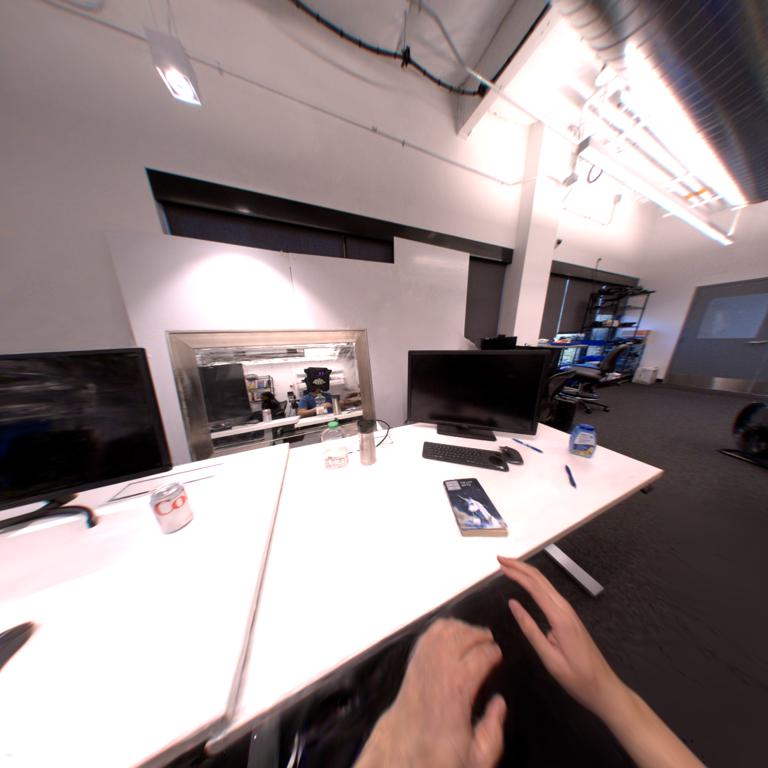}\\[-5pt]
        \includegraphics[width=0.2\textwidth]{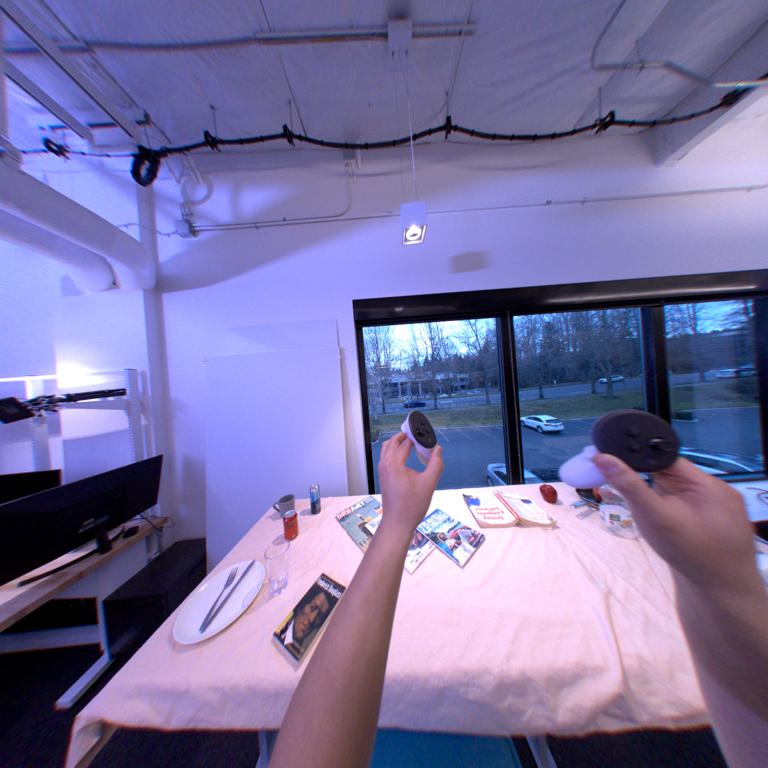} &
        \includegraphics[width=0.2\textwidth]{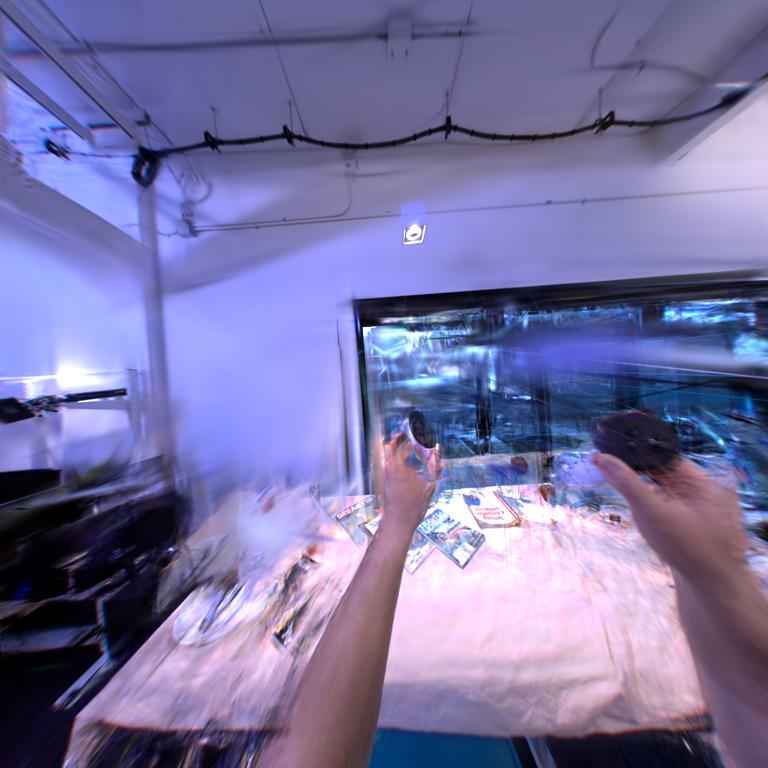} &
        \includegraphics[width=0.2\textwidth]{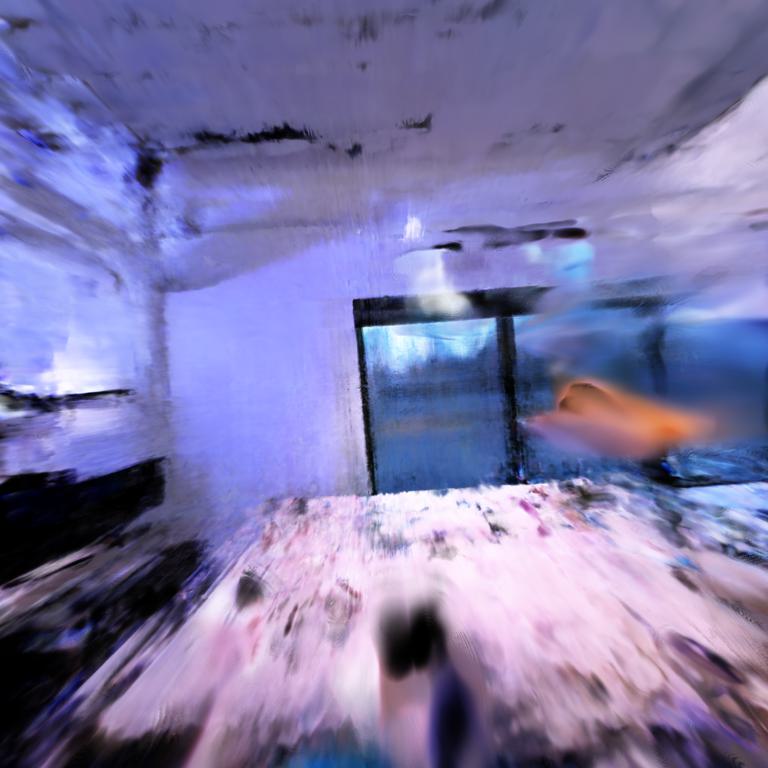} &
        \includegraphics[width=0.2\textwidth]{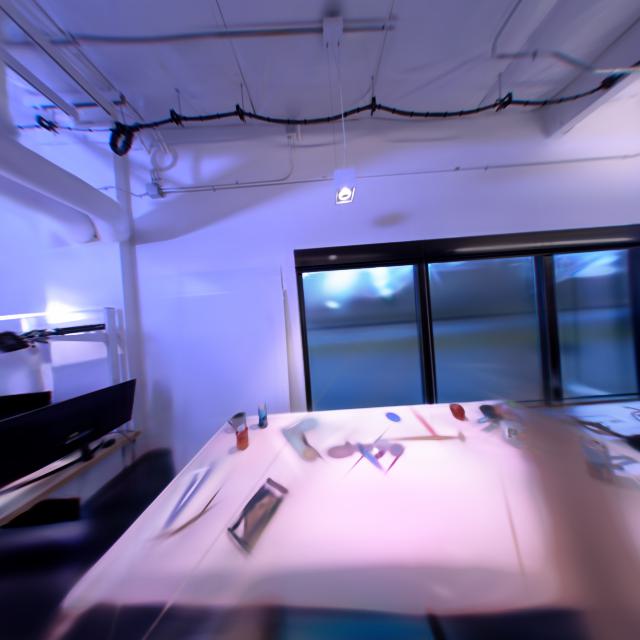} &
        \includegraphics[width=0.2\textwidth]{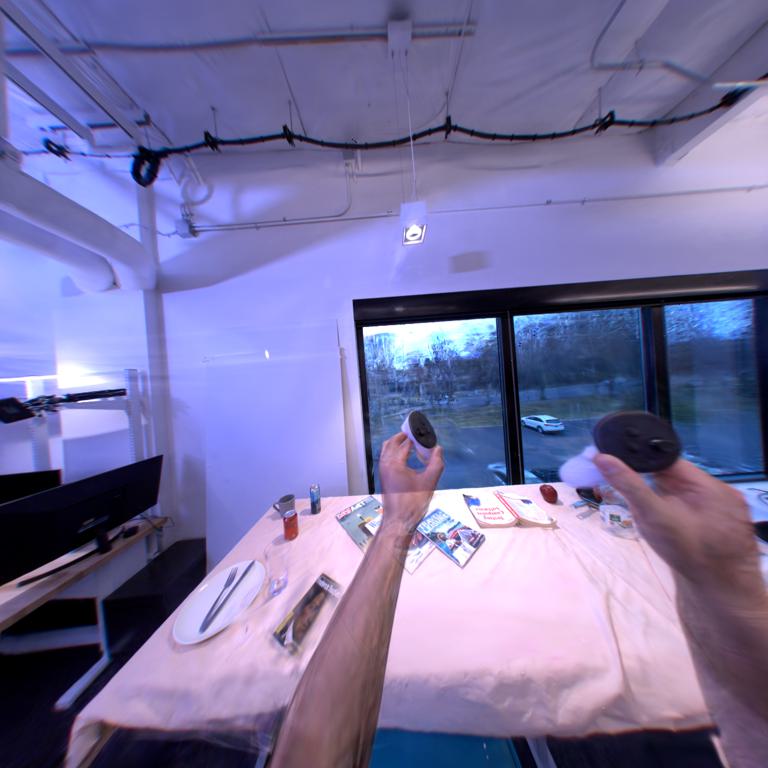}\\[-5pt]
        \includegraphics[width=0.2\textwidth]{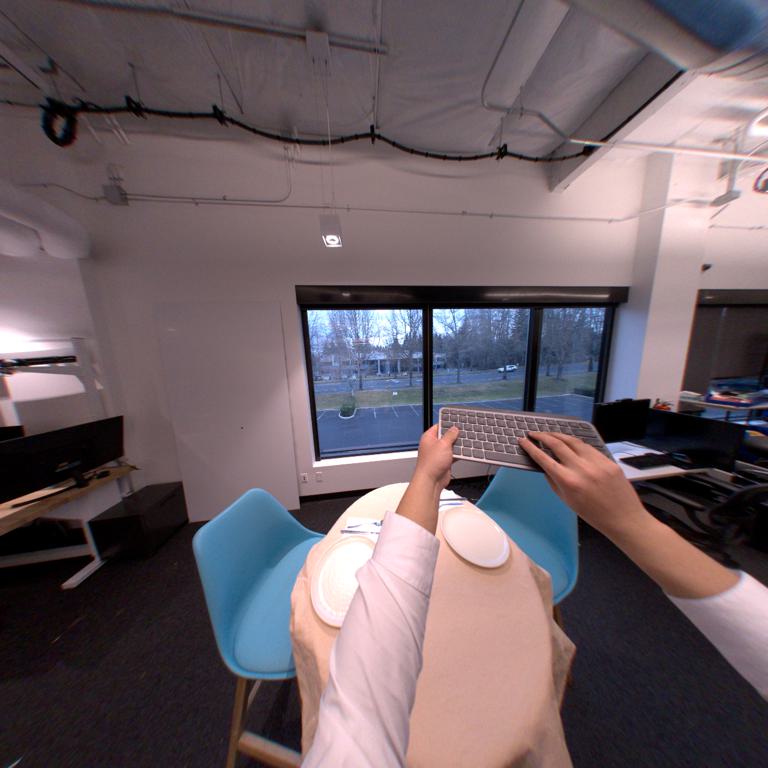} &
        \includegraphics[width=0.2\textwidth]{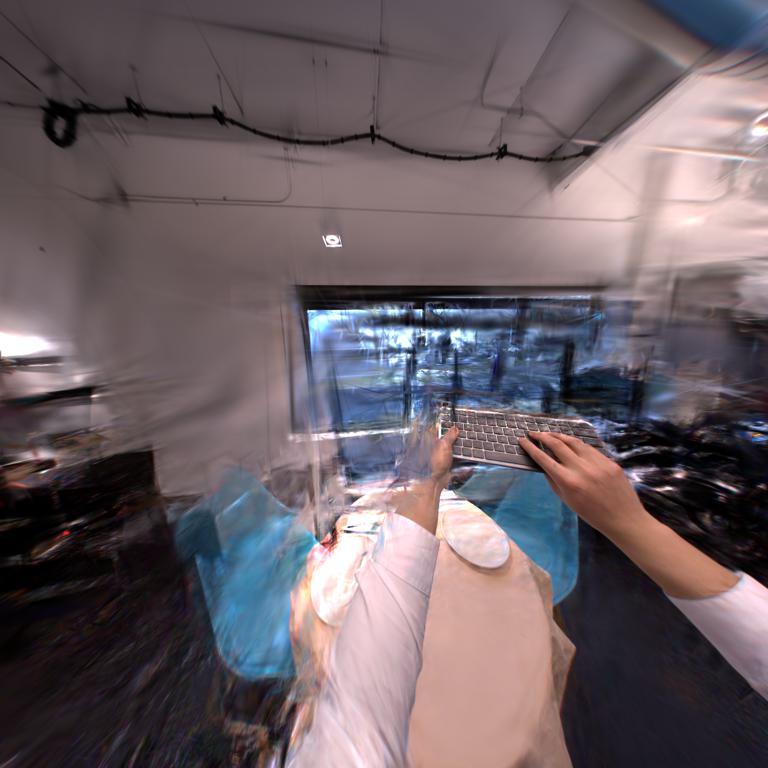} &
        \includegraphics[width=0.2\textwidth]{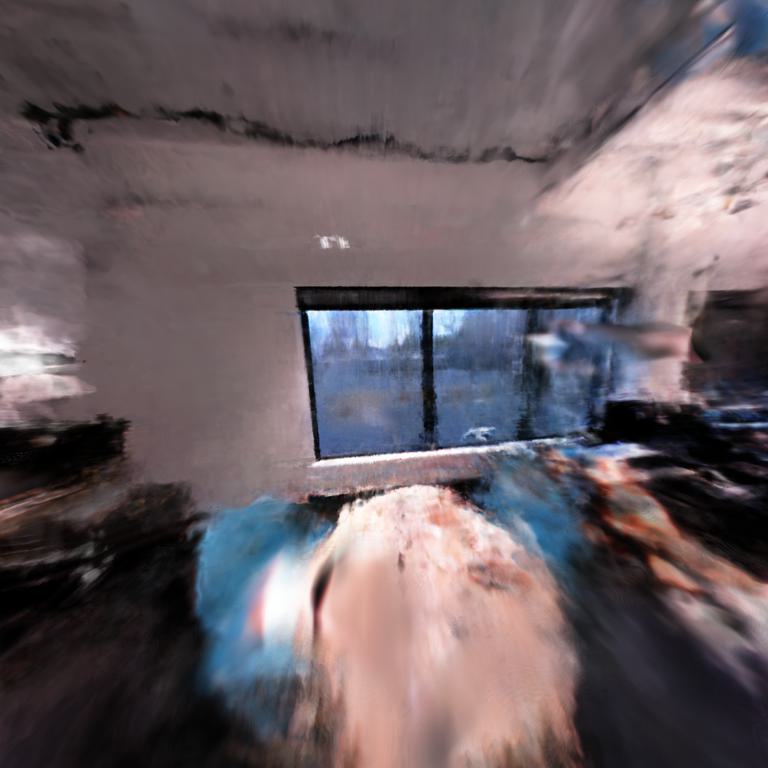} &
        \includegraphics[width=0.2\textwidth]{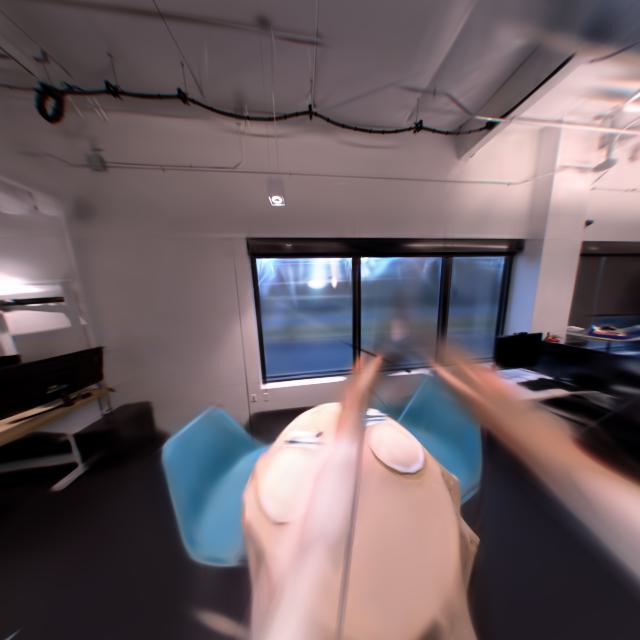} &
        \includegraphics[width=0.2\textwidth]{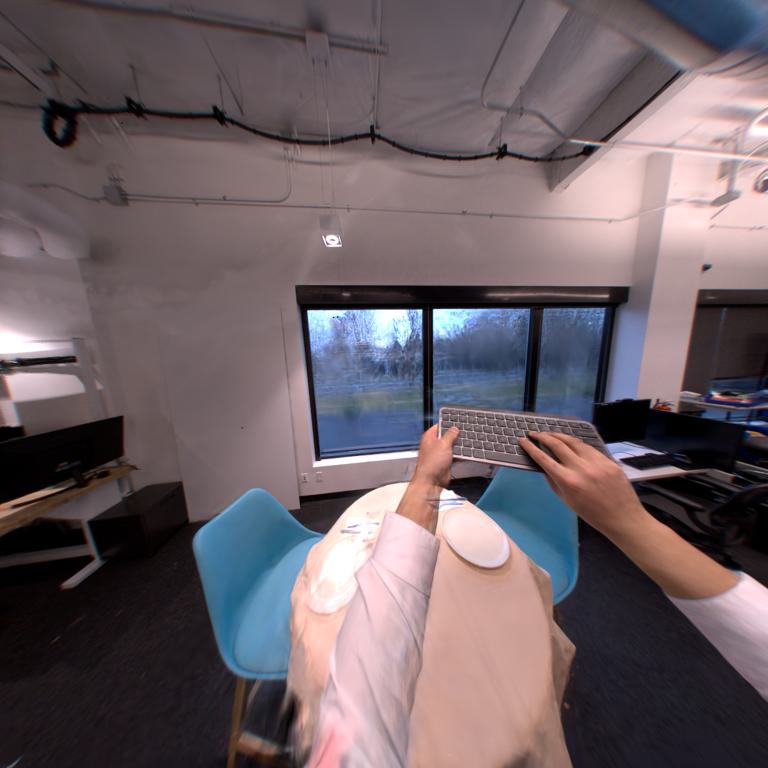}\\
        RGB Image & 3DGS & K-Planes & Spacetime Gaussians & Stereo Guided 3DGS
    \end{tabular}
    }
    \caption{
        Qualitative visualization of 4D reconstruction methods. From left to right: RGB test view and reconstructions by 
        per-frame 3DGS~\cite{gs3d_kerbl_siggraph23},
        K-Planes~\cite{kplanes_fridovich_cvpr23},
        Spacetime Gaussians~\cite{spacetime_li_cvpr24},
        and 3DGS with stereo guidance.
    }
    \label{fig:qualitative_recon}
\end{figure*}

\subsubsection*{Results}

Table~\ref{tab:nvsresults} summarizes the quantitative results of 4D reconstruction evaluation on our dataset.  We show that 3DGS with stereo guidance achieves far better results, with PSNR improvements of 7.9, 12.7 and 4.4 over original 3DGS, K-Planes, and Spacetime Gaussians, respectively. We note that the 3D Gaussian model performs better than the radiance field model, due to the large spatial extent of the scene. Figure~\ref{fig:qualitative_recon} visualizes NVS results for each fitted model. We demonstrate that fitting 3DGS per-frame is not sufficient to overcome ill-posedness, and that existing dynamic models are unable to successfully reconstruct our dataset, as they are designed to learn object-centric scenes \cite{dnerf_pumarola_cvpr21} or multiview video with fixed poses \cite{neural3dvideo_li_cvpr22, technicolor_sabatar_cvprws17}.  We find that the performance gap between our model and existing methods is wider for scenes with close dynamic objects (hands) compared to those farther away (other persons).

\begin{table}[b]
\centering
{\small
\begin{tabular}{lc@{~~}c@{~~}c} 
\toprule
Model & PSNR$\uparrow$ & SSIM$\uparrow$ & LPIPS$\downarrow$ \\
\midrule
3DGS \cite{gs3d_kerbl_siggraph23} (per-frame)  & 21.22 & 0.709 & 0.260 \\
K-Planes \cite{kplanes_fridovich_cvpr23}       & 16.46 & 0.597 & 0.443 \\
Spacetime Gaussians \cite{spacetime_li_cvpr24} & 24.76 & 0.780 & 0.270 \\
3DGS + stereo guidance                         & \bf 29.12 & \bf 0.830 & \bf 0.115 \\
\bottomrule
\end{tabular}
}
\caption{Quantitative results on 4D reconstructions. We measure average PSNR, SSIM and LPIPS over all frames for cameras in the target pair.}
\label{tab:nvsresults}
\end{table}

\section{Conclusion}
\label{sec:conclusion}

We provide Ego-1K, a large-scale time-synchronized multiview dataset, obtained with a moving egocentric rig with 12 cameras surrounding a Quest 3 headset.  This unique setup enables benchmarking of 3D video synthesis in complex, real-world dynamic environments from egocentric viewpoints, with particular focus on hand-object interaction.  To our knowledge, this is the first dataset to simultaneously achieve large scale, high camera count, egocentric perspective, and precise synchronization for dynamic scene understanding and egocentric video synthesis.

We demonstrate that current 3D and 4D NVS methods are unable to deliver accurate new views for our challenging setup involving both a moving rig and dynamic content.  In contrast, state-of-the-art foundation stereo models are able to provide decent depth maps.  Our dataset can be used to evaluate robustness to stereo baseline changes, as well as temporal stability; similar consistency studies could be performed for semantic or person segmentation.  Furthermore, we provide a baseline NVS method that uses stereo depth as guidance, which significantly improves results.  Another avenue for future work is to create high-quality per-frame depth maps via multiview / multi-baseline stereo, which could then serve as pseudo ground truth for evaluating two-view or monocular depth estimation methods, ablation studies for evaluating subsets of cameras, or evaluation of HOI-focused tasks.  We hope that our dataset will serve as a catalyst for future research along these promising lines.

\subsection*{Acknowledgments}

We thank Joey Conrad, Anton Clarkson, Pratik Halani, Daniel Ju, Max Strand, Rene van Ee, Robb Meeker, Richard McVey, Carlton Collett, and Michael Ashton for helping design and build our rig, and Aaron Ali Hawkins, Sam Coppinger, Mason Maurer, Kevin Chau, Kenneth Bradley, and Joe Park for capturing the data.

\clearpage

{
\small
\bibliographystyle{ieeenat_fullname}
\bibliography{main}
}

\end{document}